\RequirePackage{fix-cm}
\documentclass[twocolumn]{svjour3}
\smartqed  

\usepackage{hyperref}
\hypersetup{
	colorlinks=true,
	linkcolor=blue,
	filecolor=magenta,      
	urlcolor=cyan,
	citecolor=red
}

\usepackage{mathptmx}
\usepackage{amsmath}
\usepackage{amssymb}
\usepackage{epstopdf}
\usepackage{graphicx}
\usepackage{multirow}

\usepackage{microtype}

\newenvironment{centermath}
 {\begin{center}$\displaystyle}
 {$\end{center}}

\usepackage{algorithm,algpseudocode}

\newcommand{\bfJ}{\mathbf{J}}

\newcommand{\bfp}{\mathbf{p}}
\newcommand{\bfv}{\mathbf{v}}

\newcommand{\argmin}{\operatornamewithlimits{argmin}}

\journalname{Construction Robotics}

\begin{document}\sloppy

\title{Automated sequence and motion planning for robotic spatial extrusion of 3D trusses}

\author{Yijiang Huang$^1$
	\and
	Caelan R. Garrett$^2$
	\and
	Caitlin T. Mueller$^1$
}

\institute{Yijiang Huang \at
	\email{yijiangh@mit.edu}
	\and
	Caelan R. Garrett \at
	\email{caelan@csail.mit.edu}
	\and
	Caitlin T. Mueller \at
	\email{caitlinm@mit.edu} 
	\and
	$^1$Building Technology Program, Department of Architecture, Massachusetts Institute of Technology, Cambridge, MA, 02139 USA\\
	\\$^2$Computer Science and Artificial Intelligence Laboratory, Massachusetts Institute of Technology, Cambridge, MA, 02139 USA
}

\date{Received: date / Accepted: date}

\maketitle

\begin{abstract}
While robotic spatial extrusion has demonstrated a new and efficient means to fabricate 3D truss structures in architectural
scale, a major challenge remains in automatically planning extrusion sequence and robotic motion for trusses with unconstrained
topologies. This paper presents the first attempt in the field to rigorously formulate the extrusion sequence and
motion planning (SAMP) problem, using a CSP encoding. Furthermore, this research proposes a new hierarchical planning
framework to solve the extrusion SAMP problems that usually have a long planning horizon and 3D configuration complexity.
By decoupling sequence and motion planning, the planning framework is able to efficiently solve the extrusion sequence,
end-effector poses, joint configurations, and transition trajectories for spatial trusses with nonstandard topologies. This paper
also presents the first detailed computation data to reveal the runtime bottleneck on solving SAMP problems, which provides
insight and comparing baseline for future algorithmic development. Together with the algorithmic results, this paper also
presents an open-source and modularized software implementation called Choreo that is machine-agnostic. To demonstrate
the power of this algorithmic framework, three case studies, including real fabrication and simulation results, are presented.
\end{abstract}

 \section{Introduction}\label{sec:intro}
The advancement of additive manufacturing offers unprecedented
means for achieving geometrical complexity for
continuum structures at the architectural scale. However,
conventional layer-based 3D printing techniques, such as
fused deposition modeling (FDM), are not appropriate for three-dimensional discrete truss structures because of the anisotropy structural behavior incurred by the material deposition
process \cite{fang2017thesis} and prohibitively long fabrication time. In contrast, robotic spatial extrusion, sometimes called spatial 3D printing, has been proven to be an appealing alternative to fabricate bespoke 3D trusses design \cite{hack2014mesh}\cite{branch2018}\cite{tam2018}\cite{yu2016acadia}. This technique involves extruding and solidifying
thermoplastic along prescribed linear paths in space
to form spatial meshes or lattices, taking advantage of the
precision and flexibility offered by industrial robots.

However, despite the advantages of the accuracy, efficiency,
and the new fabrication possibilities offered by these
multi-axis machines, the level of automation in this designfabrication
workflow is still comparably low. While transitioning
between a digital design model and machine code
for a 3-axis gantry machine is straightforward, for multiaxis
robots, gaining fine levels of control and bypassing the
complexity of generating collision-free robotic trajectories
is much more nuanced and subtle, which requires significant
effort.

Existing investigations in the field of architectural robotics
often involve manual planning of path guidance for
the robot’s end effector, followed by tedious diagnosis for
potential problems in a trial-and-error manner. This slow
and cumbersome workflow deviates from the initial purpose
of having such a digital design-fabrication workflow:
to forge a smooth and direct transition from digital design
to real-world machine materialization; instead, the current
process often requires a complete re-program for the robot
whenever the target geometry has a small change. This
technical challenge in the sequence and motion planning
and programming of the robots congests the overall digital
design/production process and often confines designers to
geometries with standard topology with repetitive patterns.
Thus, to broaden the topological class of robotic-printable
truss structures, an automated sequence and motion planning
system is needed.

This work presents a new algorithmic framework for
robotic sequence and motion planning in the context of spatial
extrusion. The planning framework is implemented as
a flexible planning tool, called Choreo, that allows users
to input unconstrained spatial trusses and receive an automatically
generated feasible extrusion sequence and robotic
trajectories. Choreo communicates with designers in the language
of geometry and topology, while shielding the users
away from the computational complexity of sequence and
motion planning. It fills in the critical link between digital
geometry and machine fabrication, playing a similar role
of slicing algorithms in the layer-based 3D printing context.
Case studies in simulation and on real hardware are
presented to show Choreo’s power in enabling automated
planning for robotic extrusion of complex 3D trusses with
non-standard topologies, which has not been shown possible
before.
\subsection{Contributions and organization of paper}
The original contributions of this paper are summarized as: 
\begin{itemize}
\item This paper embodies the first attempt in the field of architectural
robotics to formally formulate sequence and
motion planning problem (SAMP) in the spatial extrusion
context, using the language of Constraint Satisfaction
Problem (CSP).
\item This paper presents a new hierarchical planning algorithm
to solve the SAMP problem in an integrated,
streamlined algorithmic workflow for the first time. By
rigorously formulating the problem and presenting a new
planning algorithm as a baseline, this research sets up a
corner stone for future work on the algorithmic improvement.
\item Inside the planning hierarchy, the sequence planner
solves a relaxed extrusion planning problem using a
backtracking search algorithm with user-guided decomposition
and constraint propagation; empirical results are
presented to compare the trade-off between several strategies
to assist the search and show the computational overhead
induced by the extrusion planning problem, which
differentiates it from the general discrete CSP problems.
\item The novelty of the motion planning module includes a
new sampling-based semi-constrained Cartesian planner
to enable existing motion planners to work with long
planning horizon.
\item A modularized and customizable implementation of the
proposed planning framework is presented. The planning
software, called Choreo, is adaptable to various hardware
setups and can be smoothly fitted into existing digital
design workflow.
\end{itemize}
The paper starts with a review of existing efforts in the field of robotic extrusion and planning. Section \ref{sec:assembly_planning_framework} presents the planning algorithm, starting from problem formulation(section \ref{sec:problem_formulation}) and goes through layers of its planning hierarchy: first sequence planning layer (section \ref{sec:seq_planning_module}) and then motion planning layer (section \ref{sec:motion_planning_module}). A post-processing module is presented in section \ref{sec:post_processing_module} to increase usability and adaptability of the computed trajectories. Section \ref{sec:implementation} presents the engineering ideas behind the implementation of the extrusion planning tool Choreo. Section \ref{sec:case_studies} shows three case studies with computation statistics and fabrication results to demonstrate Choreo's efficiency and power. Finally, section \ref{sec:conclusion} concludes the paper by noting out limitations and suggesting areas of future research.
 \section{Related work}\label{sec:related_work}
In response to the need for automated planning in robotic
extrusion, this section summarizes previous efforts in the
area of robotic extrusion and planning. Key research from
five distinct fields, (1) architectural robotic spatial extrusion,
(2) path planning for architectural robotic assembly,
(3) computer graphics, (4) manipulation planning, and (5)
task and motion planning is presented, with contributions
and drawbacks highlighted. The aim of this section is to
demonstrate why an integrated planning system, which
combines features from all the above fields, is needed for
robotic spatial extrusion to be fully accessible to architectural
designers.
\subsection{Architectural robotic extrusion}\label{sec:robotic_extrusion_in_arch}
Robotic extrusion (sometimes called spatial 3D printing) involves extruding a thermoplastic along linear paths, typically to form a mesh or grid
structure, using robotic motion. In recent years, this idea has been presented as an alternative to layer-based additive manufacturing for discrete spatial trusses, with advantages both in terms of mechanical properties \cite{tam2018} and speed of construction \cite{oxman2013freeform}\cite{hack2014mesh}\cite{mataerial2018}\cite{branch2018}. However, the flexibility of the industrial robots has mostly been deployed to facilitate complexity in shape (as opposed to topologies): morphed grids with repetitive zig-zag topologies have been shown to be useful both for formal variation \cite{helm2015iridescence}\cite{yuan2016robotic}\cite{soler2017generalized}\cite{branch2018} or structural efficiency \cite{tam2018}. In many among this line of work, the industrial robot follows a manually assigned zig-zag end effector path, with limited variation in the end effector's direction. Recently, Huang et al. and Yu et al. presented a constrained graph cut algorithm to tackle the extrusion sequence planning problem \cite{yu2016acadia}\cite{huang2016framefab}. Their algorithm solves for a feasible extrusion sequence along with end effector's feasible directions for each extrusion step, which enables fabrication of frames with non-repetitive topologies. However, their method abstracts away the robot's kinematics during the sequence planning process and instead uses an ad-hoc method to generate feasible guiding curves for the robot's end effector to follow. This results in slow computation and lacks any trajectory feasibility guarantees. As a step forward from their work, this work formulates the problem of combined sequence and motion planning for the first time and proposes an algorithm to solve it in an integrated, streamlined way.
\subsection{Path planning for architectural robotic assembly}\label{sec:robotic_assembly_in_arch}
Robotic spatial extrusion, although regarded as a fabrication instance, shares a similar planning problem formulation with a broader class of robotic assembly problems: given a discrete spatial structure's design model, the robot needs to be assigned a coordinated sequence of \textit{transition} and \textit{assembly} actions, to manipulate or fabricate individual elements in a specific order to construct the designated design. The solutions for these problems are alternating sequences of transition and assembly actions that corresponds the robot moving with its hand empty and while holding or extruding the element to be constructed. The assembly action usually involves constrained Cartesian movement of the robot's end effector, while transition involves free-space movement with no extra constraint other than collision-free. Specifically for robotic spatial extrusion, the assembly action corresponds to the extrusion process.

Existing exploration of robotic assembly in different architecture-scale application contexts has focused on the design of application-specific processes and associated hardware systems \cite{gramazio2014robotic}. In all these applications, researchers have encountered similar problems: (1) the assignment of an order to assemble objects and (2) the generation of feasible robotic trajectories that do not collide with objects in the workspace \cite{parascho2017cooperative}\cite{eversmann2017prefabtimber}\cite{sondergaard2016topology}. Current solutions to this problem typically involve an intuition-based trial-and-error method. For a given robot configuration during assembly, designers manually specify end-effector poses on the assembly geometry to achieve linear end-effector movement. For transition trajectories, designers manually generate guiding curves for the end-effector to follow, which hover over the workspace within a safety distance. Utilizing industrial robot's built-in commands like Linear movement (LINE) or Point-To-Point (PTP) \cite{braumann2011parametric}, users rely on the built-in interpolation method to translate end-effector assignment to joint trajectories that are free of collisions, respect joint limits and avoid singularities. As a result, this requires significant effort to diagnose the planning failure in a trial-and-error manner. Software packages exist to support this trial-and-error procedure by simulating robotic motion (such as HAL \cite{schwartz2013hal} and KUKA$|$PRC \cite{braumann2011parametric}), but these tools can only simulate/test a robotic trajectory based on Tool Center Point (TCP) planes and joint configurations input, without the ability to automatically plan a collision- and kinematics-aware trajectory. Because of this, these tools currently support a sub-optimal manual planning process.

Recent attempts at addressing this problem include partial solutions: (1) use an automated sequence planning scheme and then use either the manual planning process described above or some ad-hoc trajectory finding technique without trajectory feasibility guarantee \cite{sondergaard2016topology}\cite{yu2016acadia}\cite{huang2016framefab} (2) use a manually assigned assembly sequence and then use motion planning / online control algorithms to find feasible trajectories \cite{gandia2018towards}\cite{dorfler2016mobile}. While these approaches might be feasible for models with simple and sparse topologies, the construction sequence and robotic motion is much more nuanced for designs with denser material distribution and non-standard topologies. An integrated planning tool that combines assembly sequence and motion planning has not been developed. As pointed out in \cite{giftthaler2017mobile}, the combination of an autonomous control scheme with a ``higher-level planner'' that is ``able to negotiate cluttered environment'' is a key step to enable robotic assembly systems to operate safely in densely populated workspaces \cite{eversmann2017prefabtimber}. This work presents the first rigorous formulation of this sequence and motion planning problem, provides a new algorithm to solve it as a baseline, and shows the major computational overhead emerging in these problems. Thus, this work sets up the baseline for comparison for future development for high-level sequence and motion planners.
\subsection{Spatial extrusion in computer graphics}
In recent years, there is a growing interest in the computer graphics research community in exploring new algorithms and hardwares to enable rapid prototyping and fabrication of visual objects \cite{Baudisch_Mueller_2017}. In this line of research, the exploration on spatial 3D printing started with the work WirePrint \cite{mueller2014wireprint}, which proposes an efficient way to print wireframe meshes, where edges in the mesh are directly extruded in 3D space. A wireframe of a model is generated by slicing it horizontally and filling each slice with zig-zag filaments. This approach is limited in the types of meshes that can be printed. To improve flexibility, Peng et al. introduce a 5-DOF printer that modifies a standard delta 3D printer by adding two rotation axes to the print bed \cite{peng2016fly}. Following up this work, Wu et al. present a printing sequence planning algorithm for this 5-DOF printer \cite{wu2016printing}. Peng et al. propose a new human-machine interaction scheme using augmented reality, enabling a versatile live editing interface to link designer's modeling and robotic spatial extrusion  \cite{peng2018roma}.
\paragraph{Related problems}
In the context of 3D printing in bio-engineering, Gelber et al. presented a heuristic backtrack searching algorithm to generate printing sequence to enable micro-scale spatial 3D printing on a purpose-built isomalt 3D printer \cite{Gelber_Hurst_Bhargava_2018}. They were the first to identify that joint positioning errors are caused by beam compliance and include it as a cantilever constraint in the sequence searching process \cite{Gelber_Hurst_Bhargava_2018}\cite{Gelber_Hurst_Comi_Bhargava_2018}. This finding influenced the nodal printing orders routing part of the sequence planning module presented in this work (section \ref{sec:opt_printing_directions}).
\subsection{Manipulation planning}\label{sec:mani_planning}
The robotic planning community has developed many approaches for motion planning that identify trajectories by searching in the continuous space of robot joint angles. Recent approaches perform this search using either sampling \cite{lavalle2006planning} or optimization \cite{ratliff2009chomp}\cite{Kalakrishnan_Chitta_Theodorou_Pastor_Schaal_2011}\cite{schulman2014motion}. In manipulation planning, the goal is not only to move robot without colliding with objects in the environment, as in classical motion planning, but also to contact, operate, and interact with objects in the world. Early treatment of this problem uses a \textit{manipulation graph} to decompose planning for one robot to one object into several problems that each requires moving between connected components of the combined configuration space \cite{Alami_Laumond_Simeon_1994}\cite{simeon2004}. This work observes that solutions are alternating sequences of \textit{transit} and \textit{transfer} paths, which corresponds the robot moving with its hands empty and while holding an object. Hauser et al. identify a generalization of manipulation planning problem as \textit{multi-modal motion planning}, i.e., motion planning for systems with multiple modes, representing different sub-manifolds of the configuration space subject to different constraints \cite{Hauser_Latombe_2010}\cite{Hauser_Ng-Thow-Hing_2011}.

Rearrangement planning is a special instance of pick-and-place planning where all objects have explicit goal poses. These problems are very similar to the robotic extrusion planning problems addressed in this work, where object goal poses are specified in the input design model. Stilman et al. first introduced a version of the rearrangement problems called \textit{navigation among movable obstacles} (NAMO), where the robot must reach a specified location among a field of movable obstacles \cite{Stilman_Kuffner_2006}\cite{Stilman_Schamburek_Kuffner_Asfour_2007}. They provide a greedy backchaining algorithm for solving \textit{monotone} problem instances, where each object need only be moved onces.
Extending this work to non-monotone problem instances, Krontiris and Bekris provided an algorithm that constructs a probabilistic roadmap (PRM) \cite{Kavraki_Svestka_Latombe_Overmars_1996} in the combined configuration space, using the algorithm of Stilman et al. as connection primitive \cite{Krontiris_Bekris_2015}\cite{Krontiris_Bekris_2016}. 

Dogar et al. propose a formulation of multi-robot grasp planning as a constraint satisfaction problem (CSP) \cite{dogar2015multi}. They attempt to find short plans that requires few regrasps. However, they assume that an assembly sequence is given and does not consider reachability constraint between assembly configurations.
\subsection{Task and motion planning}\label{sec:tamp}
While motion planners deal with geometric constraints in high-dimensional configuration spaces, they do not consider abstract features of the domain, i.e. they can plan how to move the robot's joints to pick up an object but cannot decide the order of tasks to satisfy certain constraints. In contrast, the artificial intelligence (AI) planning community considers problems that are discrete but require many types of actions to be performed over long horizons without predefined horizon length \cite{mcdermott1998pddl}\cite{helmert2006fast}. Recent work in task and motion planning (TAMP) \cite{Dornhege2012}\cite{Toussaint_2015}\cite{lagriffoul2016combining}\cite{garrett2016ffrob} combines AI and motion planning to simultaneously plan for discrete objectives as well as robot motions. This work aims to enable robots to operate in applications such as cooking, which require discrete choices of which objects to grasp or cook as well as continuous choices of which joint angles and object poses can physically perform each task. A key challenge is that often physical constraints such as collision, kinematic, and visibility constraints can restrict which high-level actions are feasible. Readers are referred to \cite{Garrett_Lozano-Perez_Kaelbling_2018} for a more complete review of the work in this area.

Lagriffoul et al. propose a constraint-satisfaction approach to interleave the symbolic and geometric searches. They focus on limiting the amount of geometric backtracking \cite{lagriffoul2014efficiently}. Lozano-P\'erez and Kaelbling take a similar approach but leverage CSPs operating on discretized variable domains to bind free variables \cite{lozano2014constraint}. The sequence planning module (section \ref{sec:seq_planning_module}) proposed in this work adopts a similar technique by using CSP to bind free geometric variables on a plan skeleton. However, it relaxes the requirement on feasible whole paths' existence and trades the algorithm's completeness for scalability.

\subsection{Sequence and motion planning for spatial extrusion: unique challenges and unmet needs}\label{sec:assembly_planning_review}

The sequence and motion planning (SAMP) problem in spatial extrusion context is a subclass of high-dimensional robot manipulation problems, or more generally, task and motion planning (TAMP) problems, which require planning a coordinated sequence of motions that involve extrusion as well as moving through free space. The SAMP problems involve two key aspects that differ them from typically studied TAMP problems: (1) fixed but long planning horizon, (2) physical constraints that involve computation overhead.

First, the discrete horizon of the SAMP problems is much longer than many TAMP benchmarks \cite{lagriffoulbenchmarks}, which often only require manipulating a couple of objects. Because each element must be extruded once and the goal object poses are specified by the input design geometry, the planning horizon equals to the number of elements in the input model, which is known in advance. Thus, SAMP requires identifying an order for extruding linear elements, fitting this order to a fixed-length plan skeleton, and binding the required geometric parameters. In contrast, TAMP problems generally have unsettled action plans - it is not initially clear which actions are needed and in which order to perform these actions to complete a task and thus, the planning horizon can be arbitrarily long. 

Second, SAMP problems involve physical constraints such as stiffness and stability that are not typically found in TAMP benchmarks. These constraints impact many state variables at once, making them challenging to effectively incorporate in many discrete task planning algorithms. Rather than directly using existing TAMP algorithms, a specialized system is developed in this research that incorporates several existing ideas but, because of its specialization to assembly planning, can scale to complex models. 

Apart from the planning problems' characteristics, common task specification languages for discrete AI planning systems, such as Planning Domain Definition Language (PDDL) \cite{mcdermott1998pddl} are not intuitive for architects and designers. The requirement of specifying task domains, predicates, action's preconditions and effects departs from the architectural language of shape and geometry, and thus creates a gap between an architect's geometric model and robotic task specification for planning. This gap in the modeling interface inhibits these algorithms from being easily adapted to architectural robotic assembly applications.

In summary, there is a rich literature of work related to robotic spatial extrusion for architecture, ranging from theoretical research in robotic task and motion planning to examples of built work of considerable intricacy. However, the field is nevertheless lacking an integrated, general-purpose method that can be applied systematically to handle the geometric and topological complexity of 3D trusses in the context of contemporary architectural design.  This work addresses this gap by presenting a new sequence and motion planning algorithm framework and a modularized implementation. Although the algorithm described in this paper is more specialized than most of the TAMP approaches, the ability to scale to problems with much longer planning horizon and a larger branching factor is one of the key focuses of this research.
 \section{Sequence and motion planning for robotic extrusion}\label{sec:assembly_planning_framework}
This section first presents a new problem formulation for the sequence and motion planning (SAMP) problem involved in robotic extrusion (section  \ref{sec:problem_formulation}). Then, a new algorithmic framework is presented to solve the SAMP problem. Section \ref{sec:concept_overview} gives a conceptual overview of the framework's hierarchy and introduces its three main modules. Detailed descriptions are then stated for the \textit{sequence planning module} (section \ref{sec:seq_planning_module}), the \textit{motion planning module} (section \ref{sec:motion_planning_module}) and the \textit{post-processing module} (section \ref{sec:post_processing_module}).
\paragraph{Assumptions}
In this paper, the robot is assumed to work in a fully observable and deterministic environment. The generated plan is purely geometric - the computed velocities are not used in the execution. Trajectory's speed for execution is reassigned separately by the user after the planning is finished and robot's position control is carried out by the industrial robot's controller.
\paragraph{Model input}
The extrusion planning problem starts with a 3D truss model from a designer. The 3D truss is represented via a standard node-member data representation. Nodes are described with 3D spatial coordinates in an indexed list. Linear members are described by their start and end node indices. A uniform cross section and material property are assumed for all of the elements. Each linear element specifies a sequence of path points that the end effector's tip must extrude along. 

\subsection{Planning problem formulation}\label{sec:problem_formulation}
The extrusion sequence planning problem has a predefined plan skeleton, or action sequence, that has an alternating pattern on the actions: {\it extrude(element $O_i$) - transit - extrude(element $O_{i+1}$) - transit}. The planner needs to assign a correct assignment of object $O_i$ to each action in the plan skeleton, and bind variables to fully specify robot's configurations \textit{during} and \textit{between} extrusion steps, where different constraints exist for the robot's motion in the two types of actions. In the following discussions, let $n$ denote the total number of elements to be extruded in the model.

\subsubsection{Free-space joint motion and semi-constrained Cartesian motion}\label{free_cart_motion}
%
The robot's motion is subject to different constraints during transition and extrusion process in the action sequence. The transition process involves a segment of the robot's trajectory after it finishes extruding the last element, and before it starts extruding the next one. Thus, this connecting trajectory requires only a collision-free joint trajectory with no extra requirement on the position and the orientation of the end effector, i.e. \textit{free-space joint motion}. 

\begin{figure}[!htbp]
 \includegraphics[width=0.25\textwidth]{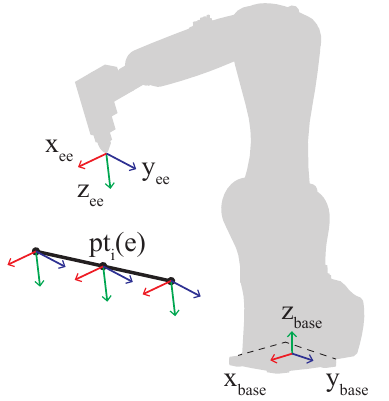}
 \centering
 \caption{Path points ${pt_{i}(e)}$ are specified by discretizing element $e$, described in world frame $\{x_{base},y_{base},z_{base}\}$. The end effector (EE)'s behavior for extruding $e$ is defined by assigning an end effector's frame $\{x_{ee}, y_{ee}, z_{ee}\}$ and translating along the path points ${pt_{i}(e)}$.}
 \label{fig:eef_definition}
\end{figure}

\begin{figure}[!htbp]
 \includegraphics[width=0.3\textwidth]{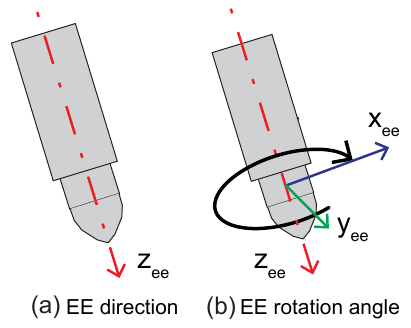}
 \centering
\caption{The end effector's orientation can be defined by a directional vector (a) and a rotation angle (b).}
\label{fig::EE_dir_rot}
\end{figure}

In contrast, extrusion process involves having the robot's end effector (EE)'s tip linearly traverse the path points that a target element defines with some fixed orientation. The EE (a 3D printing extruder) is required to maintain its orientation when extruding to obtain a straight printing result, which avoids the twisting force that the EE might exert on the molten plastic beam during extrusion \cite{Gelber_Hurst_Comi_Bhargava_2018}. Formally, an EE's orientation is defined by a frame $\{x_{ee}, y_{ee}, z_{ee}\}$ (figure \ref{fig:eef_definition}). The definition of an orientation can be alternatively described by a \textit{direction vector} and a \textit{rotational angle}, where the direction vector represents the z axis $z_{ee}$ and the rotation angle defines the x-y plane $\{x_{ee}, y_{ee}\}$ by rotating around $z_{ee}$ (figure \ref{fig::EE_dir_rot}). The \textit{pose} of an EE is specified by an affine transformation that includes a 3D position and an orientation, i.e. the commonly used TCP plane. Using this terminology, the extrusion process involves a semi-constrained Cartesian motion on the EE poses, where the extrusion element itself only specifies EE's positions but leaving two degrees of freedom on the EE direction and rotation angle undetermined. Notice that given a set of EE poses, an extra step of inverse kinematics computation and joint solution selection needs to be carried out to fully specify the robot's joint configurations during extrusion.



%

%
\subsubsection{Constraint variables}
In this section, a constraint satisfaction problem (CSP) encoding of the extrusion SAMP problem is presented. Each action in the predefined plan skeleton is specified with a symbolic constraint variable and a set of geometric constraint variables. The involved variables are of the following three types:

\begin{itemize}
\item{$O_i­, \,i \in \{1, \dots, n\}$: The element assignment for the $i$-th extrusion action. Its value domain is $\{1, \dots, n\}$, representing the indices of elements in the input model.}
\item{$H_i, \,i \in \{1, \dots, n\}$: The end effector's orientation for the $i$-th extrusion action. As described in section \ref{free_cart_motion}, the domain of $H_i$ can be discretely parameterized by a direction vector $\mathbf{v}$ on a unit sphere and a rotation angle $r \in [0, 2\pi)$.}
\item{$K_{i,j}, \,i \in \{1, \dots, n\}, \,j \in \{1, \dots, E_i^d\},$}: The robot's configuration with its end effector extruding at $i$-th extrusion element's $j$-th path point. The domain of $K_{i,j}$ is a set of joint values of a robot. $E_i^d$ denotes the path point discretization number of element $i$.
\end{itemize}

A extrusion plan skeleton is \textit{realized} if all of the constraint variables are binded to concrete values. To bind these variables, a CSP planner is called to verify if the plan skeleton is satisfiable. The correctness of a plan skeleton is enforced by the constraints, which are expressed as relationships between the constraint variables. 
Constraints are specified by a set of variables to which they apply and a test predicate that maps an assignment of variable values to true or false \cite{dechter2003constraint}.



%
%
\paragraph{Constraints}
Constraints relate the variables to one another and limit the set of valid assignments. If all the constraints are collectively satisfiable, then a plan skeleton is valid, and the pruned geometric variable domains specify the geometric details for EE's poses during extrusion. The CSP formulation includes the following constraints:

\textproc{AllDiff}($O_1, \dots, O_n$): Each element is used only once by an extrusion action.  Thus, all extrusion element assignment $O_i$'s values are distinct. This constraint is equivalent to enforce that the resulting assignment $\{O_i\}$ is a permutation of $\{1, \dots, n\}$.

\textproc{Connect}($O_1, \dots, O_k$), $k = 1, \dots, n$: At each extrusion step, the newly added element must be either connected to the existing structure or connected to the ground. Let Boolean matrix $A \in \{0,1\}^{m \times m}$ denotes the adjacency matrix of the input spatial truss design model:
\begin{centermath}
  A_{i,j} =
  \left\{
  \begin{array}{ll} 1, &\textrm{if element $O_i$ and $O_j$ share a node;}\\ 
  0, &\textrm{otherwise.}
  \end{array}
  \right.
\end{centermath}
And ground connectivity matrix $G \in \{0,1\}^{m \times 1}$:
\begin{centermath}
  G_{i} =\left\{
  \begin{array}{ll} 1, &\textrm{if element $O_i$ has a grounded node.}\\ 
  0, &\textrm{otherwise.}
  \end{array}\right.
\end{centermath}
Then the connectivity constraint can be expressed as:
\begin{centermath}
\begin{aligned}
\forall \, 1 \leqslant i \leqslant m, \, \exists \, 1 \leqslant j < i,\\
A_{O_i,O_j} = 1 \quad OR \quad G_{O_i} = 1
\end{aligned}
\end{centermath}



\textproc{Stiffness}($O_1, \dots, O_k$), $k = 1, \dots, n$: The stiffness constraint ensures that the partially extruded structure is stiff at each assembly step and the maximal deformation due to gravity (or other constantly presented load) is bounded by a predefined tolerance. The deformation of all the nodes under gravity can be calculated using finite element analysis of linear frame structures \cite{McGuire_Gallagher_Ziemian_1999}. The constraint test function returns true if and only if the maximal node deformation is smaller than a tolerance.

\textproc{Stability}($O_1, \dots, O_k$), $k = 1, \dots, n$: The stability constraint checker returns true if the gravitational center's projection on the supporting plane lies in the convex hull of all the grounded nodes, and returns false otherwise. It guarantees that the rigid, partially-extruded structure meets moment equilibrium and does not require a tension connection at the support to remain upright \cite{Phear1850}. This calculation can be easily integrated with the stiffness constraint's computation, by evaluating if there exists a tensile force among the reaction force in the grounded elements.

\textproc{ExistValidFreeSpaceMotion}($O_{static}, O_1, \dots, O_k, \\ K_{O_{k-1}, E_{O_{k-1}}^d}, K_{O_k, 1}$), $k = 2, \dots, n$:  Viewing already extruded elements $O_1, \dots, O_{k-1}$, together with static scene objects $O_{static}$, as collision objects, this constraint checks if there exists a free space joint motion to connect from last robot's configuration $K_{O_{k-1}, E_{O_{k-1}}^d}$ for extruding element $O_{k-1}$ and the first robot's configuration $K_{O_k, 1}$ for extruding the next element $O_{k}$.

\textproc{ExistValidExtrusionMotion}($O_{static}, O_1, \dots, O_k, \\ H_{O_k}, K_{O_k,1}, \dots, K_{O_k,E_{O_{k}}^d}$), $k = 1, \dots, n$: 
This constraint tests if there exists a collision-free extrusion motion to have the robot's EE linearly traverse the path points $pt_{j}(O_k), j=1,\dots,E_{O_{k}}^d$ defined by the element $O_k$ with EE orientation $H_k$ (section \ref{free_cart_motion}). The compatibility between the robot's configurations and the EE poses is enforced by checking  \textproc{InverseKinematics}($H_{O_k}@pt_{j}(O_k)) = K_{O_k,j}, j=1,\dots,E_{O_{k}}^d$. Elements $O_1, \dots, O_{k-1}$ and static objects $O_{static}$ are considered as collision objects and checked against robot's configurations.

Notice that there are three main features that makes the CSP problem here hard to solve: (1) there are a large number of robot configuration variable $K_{i,j}$ to be binded. (2) The  geometric constraint variable, $H_i$ and $K_{i,j}$, requires a discretization of the end effector pose space and the robot's joint space. The discretization granularity is a meta-parameter that is essential to balance the computation's completeness and tractability. (3) The \textproc{Stiffness} and \textproc{Stability} involves physical simulation, while \textproc{ExistValidFreeSpaceMotion} and \textproc{ExistValidExtrusionMotion} involves calling single-query motion planner or kinematics solver. The evaluation of all these constraints commonly induces a large amount of overhead as they will be called many times by the CSP planner, and they impact many variables at the same time.

To get around this computational complexity, a specialized hierarchical planning algorithm is proposed to break this problem into isolated, simplified subproblems. The algorithm trades its completeness for tractability, especially to harness the long planning horizon induced by complex design inputs.

%

%
\subsection{Conceptual overview of the planning framework}\label{sec:concept_overview}



\begin{figure}
\includegraphics{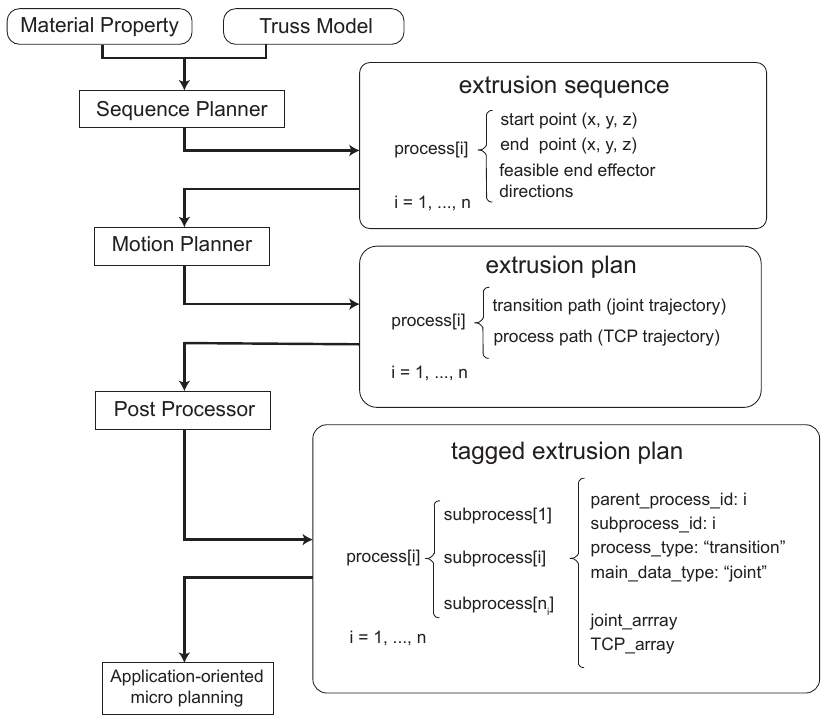}
\centering
\caption{Overview of the sequence and motion planning framework.}
\label{fig:system_overview}
\end{figure}

The proposed hierarchical planning algorithm incorporates three key modules as shown in figure \ref{fig:system_overview}. Instead of searching for a solution considering all parts of the searching tree at once, the proposed approach identifies and breaks the problem into two isolated sub-problems, sequence planning and motion planning. Specifically, the satisfaction of the \textproc{ExistValidFreeSpaceMotion} is relaxed in the sequence planner, and the determination of transition motion is postponed until the motion planner. This separation cuts the sequence-dependent ties between the sequence and motion planning subproblems, narrowing down the search space. First, the sequence planner (section \ref{sec:seq_planning_module}) generates an extrusion sequence and associated feasible end-effector directions. Next, given the fixed assembly sequence and a focused set of end-effector directions, the motion planner (section \ref{sec:motion_planning_module}) chooses the end-effector pose for each element's extrusion and plans for the robot's entire joint trajectory during and between extrusions. Finally, the post processor (section \ref{sec:post_processing_module}) tags the computed trajectory plan with associated assembly information and outputs a complete extrusion plan. After this is completed, the user can optionally use the post processor's tagging system to insert tool path modification or control commands to fine-tune the hardware control.
These modules, along with the framework inputs and outputs, are described in greater detail in the following sections. An example problem of using a fixed-base 6-axis robot is used to illustrate the details of each module, but the system is general and can also apply to other robotic extrusion systems.
\subsection{Sequence planning module}\label{sec:seq_planning_module}
The proposed sequence planner takes a 3D truss model as input and solves for the order of the extrusion operation and associated feasible end-effector directions. The sequence planner solves a relaxed version of the original CSP problem specified in section \ref{sec:problem_formulation} with a standard backtracking search strategy. The backtracking search is guided by an application-specific variable ordering, forward checking, and constraint propagation. The novel contribution of this sequence planning module are (1) the relaxed CSP formulation (2) some empirical computation statistics from a straightforward guided backtracking search solver, to demonstrate the computational bottleneck and establish the first baseline for future algorithmic development and comparison.


\subsubsection{Relaxed CSP constraints}\label{sec:seq_planning_problem_formulation}
In the original CSP encoding in section \ref{sec:problem_formulation}, the robot configuration variables $K_{i,j}$ tie the two constraint \textproc{ExistValidFreeSpaceMotion} and \textproc{ExistValidExtrusionMotion} together, inducing large computational overhead. To avoid this, a relaxed CSP formulation is proposed to eliminate the robot configuration variables and simplify the motion constraints.

\begin{figure}[!htbp]
 \includegraphics[width=0.35\textwidth]{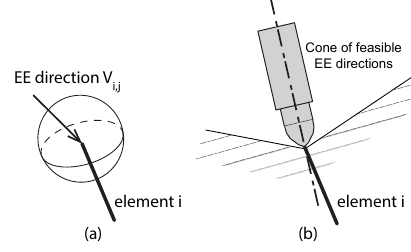}
 \centering
\caption{(a) EE direction constraint variables $V_{i,j}$ corresponds to a list of directional vectors uniformly sampled from the unit sphere. The sampling granularity $m_V$ is specified by the user. (b) Because of the existence of existing elements $O_{i}$ and static objects, the collision-free EE directions are usually limited within a cone.}
\label{fig::EE_dir_pruning}
\end{figure}

\paragraph{Constraint variables} Apart from eliminating the robot configuration variables $K_{i,j}$, the EE orientation variables $H_i$ is replaced by $V_i$, which represents the EE direction. The updated constraint variables are:

\begin{itemize}
\item{$O_i­, \,i \in \{1, \dots, n\}$: The element assignment for the $i$-th extrusion action as before.}
\item{$V_{i,j}, \,i \in \{1, \dots, n\}, \,j \in \{1, \dots, m_{V}\}$: The end effector direction's feasibility for the $i$-th extrusion action. The index $j$ refers to an ordered list of unit vectors uniformly sampled on a unit sphere (figure \ref{fig::EE_dir_pruning} (a)). The size of this sampled vector list is $m_{V}$. The value domain of $V_{i,j}$ is $\{0, 1\}$, which represents the direction is feasible (0) or not (1). }
\end{itemize}

\paragraph{Constraints}
The constraints \textproc{AllDiff}, \textproc{Connect}, \textproc{Stiffness} and \textproc{Stability} are kept in the relaxed CSP formulation. The motion-related constraints \textproc{ExistValidFreeSpaceMotion} and \textproc{ExistValidExtrusionMotion} are replaced with constraints on EE directions, which are simplified and easier-to-check:

%

\textproc{ExistValidExtrusionEEPose($O_{static}, O_1, \dots, O_k,\\V_{O_k,1}, \dots, V_{O_k,m_V}$), $k = 1, \dots, n$:} \\This constraint checks if there exists one valid end effector direction for extruding element $O_k$. Existing elements $O_1, \dots, O_{k-1}$and $O_{static}$ are considered as collision objects. These collision objects may collide with the end effector if extruding with some specific EE directions (figure \ref{fig::EE_dir_pruning} (b)).

This constraint can be expressed as:
\begin{centermath}
\begin{aligned}
&\exists \, a, 1 \leqslant  a \leqslant  m_V,\\
&\forall \, 1 \leqslant  j < i,\\
&T_{O_{k},O_{j}, a} = 1 \quad AND \quad V_{O_{k}, a} = 0\\
&AND\\
&\textproc{ExistValidKinematics}(a, O_1, \dots, O_{k-1}, O_{static}, a)
\end{aligned}
\end{centermath}
where $T \in \{0,1\}^{n \times n \times m}$:
\begin{centermath}
  T_{i,j,a} =\left\{
  \begin{array}{ll} 1, &\textrm{if the EE does not collide with element $j$}\\
  &\textrm{while extruding element $i$ with direction $V_a$}\\
  0, &\textrm{otherwise.}
  \end{array}\right.
\end{centermath}

The subfunction \textproc{ExistValidKinematics} checks if there exists valid robot kinematic solutions to extrude along element $O_k$'s path points with EE direction $V_{a}$. The constraint checker samples a rotation angle $r \in [0, 2\pi)$, which fully determines a list of EE poses during the extrusion of element $O_k$ when combined with direction $V_{a}$ and path points $pt_i(O_k)$. The subfunction checker returns true if there exist a collision-free inverse kinematics solution for each one of the EE poses and return false if no solution is found within a user-specified timeout.

Notice that the checking of $T_{i,j,a}$'s value involves only considering the EE's solid body and other collision objects, while \textproc{ExistValidKinematics} involves several IK computation and collision detection between the robot's fully-body configuration and other collision objects.

%
\subsubsection{Solving the CSP}\label{sec:solving_CSP}
A key advantage of a CSP formulation is that if a user provides a description of their problem in this representation, a generic solver can perform the search. In this paper, a simple backtracking search with value ordering and constraint propagation is used as a baseline solver. A domain-dependent heuristic is proposed to assist the value ordering. In addition, to limit the computation in a reasonable amount of time, a user-guided model decomposition is introduced before running the search algorithm. A winning strategy of only incorporating constraint propagation and model decomposition is selected based on numerical runtime results.

While integrating the CSP encoding with generic, blackbox CSP solvers is left as future work, the key contribution of this section is to establish a baseline for future algorithmic development to compare against. By providing detailed computation statistics, computation bottleneck on checking physical and geometrical constraints can be observed that distinguishes extrusion planning problems from general discrete CSP problems.
\paragraph{Backtracking search}
The proposed backtracking search algorithm is described in Algs. \ref{alg:bt_search}. It is identical to a generic backtracking search algorithm as presented in a standard textbook (chap. 6.3, \cite{russell2016artificial}), except the following parts:

\begin{itemize}
\item{The algorithm selects unassigned variable by simply advancing along the time index $k$. It only searches on the extrusion object assignment variable $O_k$ and leaves EE direction variables $V_{i,j}$ pruned by constraint propagation. Thus, the maximal depth of the search tree is $n$.}
\item{A domain-dependent heuristic is used to guide the value ordering (line \ref{line:value_order}, Algs. \ref{alg:bt_search}).}
\item{The \textproc{TestConsistency} function (line \ref{line:test}, Algs. \ref{alg:bt_search}) involves checking physical and geometrical computation, which might creates computational overhead.}
\item{The constraint propagation (line \ref{line:propagate}, Algs. \ref{alg:bt_search}) assumes the existence of element $O_k$ and updates the EE direction $V_{O_i,j}$'s domain for all the unassigned elements $O_i$. This invokes a large amount of collision checking between the element $O_k$ and the EE extruding along future element $O_i$ with directions that are still in $V_{O_i,j}$'s feasible domain.}
\end{itemize}

\paragraph{Value ordering}
The dynamic value ordering function \textproc{OrderValues} uses a \textit{collision cost} to order the value candidates. Although the element $O_k$ at current step is printable, it might cause the remaining unprinted elements to have no feasible end effector direction in the following stage. Thus, this collision cost is added for tie-breaking by prioritizing the successor that roughly admits the most future orientations. Notice that the computation of this cost involves constraint propagation on EE direction states and all the candidate value for the variable $O_k$ in consideration, which might invokes computation overhead. The impact of the collision cost on computation time is discussed in the later paragraph on Backtracking search computation statistics.

%
%
\begin{algorithm}
\newcommand{\vars}{\texttt}
\newcommand{\func}{\textproc}
\let\oldReturn\Return
\renewcommand{\Return}{\State\oldReturn}
\caption{Backtracking search}
\label{alg:bt_search}
\begin{algorithmic}[1]
\Function{BacktrackSearch}{$\vars{k}, \vars{assignment}, \vars{csp}$}
\If{\vars{$O_{n}$} is assigned} \label{line:term}
\Return \vars{assignment}
\EndIf
\State $\vars{k} = \vars{k}+1$
\State \vars{var} = $O_k$
\For{\textbf{each} $\vars{val}$ in \func{OrderValues}($\vars{var}, \vars{assignment}, \vars{csp}$)} \label{line:value_order}
\If{\func{TestConsistency}(\vars{val} , \vars{assignment})} \label{line:test}
\State add \{ \vars{var} = \vars{val} \} to \vars{assignment}
\State \vars{ee-inferences} = \func{UpdateEEDirectionState}(\vars{$O_k$}, \vars{assignment}) \label{line:propagate}
\State \vars{result} = \func{BacktrackSearch}($\vars{k}, \vars{assignment}, \vars{csp}$)
	\If{\vars{result} $\neq$ \vars{failure}}
		\Return \vars{result}
	\EndIf
\EndIf
\State Remove {\vars{var}=\vars{val}} and \vars{ee-inferences} from \vars{assignment}
\EndFor
\Return \vars{failure}
\EndFunction
\end{algorithmic}
\end{algorithm}

\begin{table*}[h]
\centering
\newcommand{\tabincell}[2]{\begin{tabular}{@{}#1@{}}#2\end{tabular}}
\newcommand{\crossmark}{$\times$}
\begin{tabular}{| c | c | c | c | c | c | c | c | c | c | c | c | c | c |}
\hline
 model &$|E|$ & \tabincell{c}{decomp.} & \tabincell{c}{collision\\cost} &\tabincell{c}{total\\time[s]} &\tabincell{c}{partial\\ states\\visited} & \multicolumn{2}{c|}{\tabincell{c}{test stiffness\\ \& stability\\(time[s]$|$cnt)}} & \multicolumn{2}{c|}{\tabincell{c}{test\\kinematics\\(time[s]$|$cnt)}} & \multicolumn{2}{c|}{\tabincell{c}{update EE \\ direction state\\(time[s]$|$cnt)}} &\multicolumn{2}{c|}{\tabincell{c}{computing\\collision cost\\(time[s]$|$cnt)}}\\
 \hline
 \multirow{2}{*}{fig\ref{fig::csp_st}(a)}
 &\multirow{2}{*}{23}
 &\multirow{2}{*}{\crossmark}
 & \crossmark & 9.8 & 39 & 0.72 & 261 & 6.38 & 273 & 2.7  & 39 & - & -\\
  \cline{4-14}
 & & & \checkmark & 18.94 & 23 & 0.36 & 150 & 0.94 & 160 & 2.64 & 23 & 14.78 & 160\\
 \hline
  \multirow{4}{*}{fig\ref{fig::csp_st}(b)}
 &\multirow{4}{*}{84}
 &\crossmark &\crossmark & $>6$hrs & - & - & - & - & - & -  & - & - & -\\
 \cline{3-14}
 & &\checkmark &\crossmark & 64.43 & 90 & 5.67 & 481 & 17.40 & 621 & 41.11 & 90 & - & -\\
 \cline{3-14}
  & &\crossmark &\checkmark & 580.35 & 84 & 8.93 & 1402 & 9.87 & 1538 & 30.65 & 84 & 528.19 & 1528\\
  \cline{3-14}
   & &\checkmark &\checkmark & 79.34 & 84 & 4.01 & 446 & 4.21 & 582 & 30.52 & 84 & 40.40 & 581\\
 \hline
  \multirow{4}{*}{fig\ref{fig::csp_st}(c)}
 &\multirow{4}{*}{132}
 &\crossmark &\crossmark & $>6$hrs & - & - & - & - & - & -  & - & - & -\\
 \cline{3-14}
 & &\checkmark &\crossmark & 103.82 & 142 & 9.54 & 715 & 14.74 & 928 & 79.35 & 142 & - & -\\
 \cline{3-14}
  & &\crossmark &\checkmark & 1871.38 & 136 & 34.49 & 3099 & 31.63 & 3313 & 77.04 & 136 & 1710 & 3262\\
  \cline{3-14}
   & &\checkmark &\checkmark & 147.42 & 132 & 9.80 & 704 & 8.10 & 914 & 75.34 & 132 & 53.54 & 890\\
 \hline
\end{tabular}
	\caption{Backtracking search algorithm runtime statistics. cnt stands for count. The discretization of directional vectors sampled on a unit sphere: 72, kinematics sampling timeout: 2s.}
	\label{table:CSP_overhead}
\end{table*}
\paragraph{User-guided model decomposition}
Model decomposition involves grouping the discrete input model into several connected components. Taking advantage of a user's intuition on the geometric relationship, the decomposition breaks the whole sequencing problem into several smaller ones, and the search is confined to each of these small sub-problems. This decreases the size of the search space and leads to more efficient CSP solving overall. When the input model has a large number of elements, the runtime caused by computing the collision cost increases drastically, as it is equivalent to call \textproc{UpdateEEDirectionState} everytime a cost is computed. The model decomposition limits the states considered by the constraint propagation.



However, it is possible for the user to provide a bad decomposition that leads to longer runtime or even failure of finding a feasible solution. Based on the authors' experiments, it usually takes several iterations before one can find one decomposition that works. Nevertheless, searching with decomposition provides users a quicker way to have a sense of whether the model has a sequence solution or not, while a direct heuristic search can only assert the inexistence of a solution after searching all possible partial states. A more general automatic model decomposition, along with the relationship between decomposition and completeness, is left as future work.
\begin{figure}
 \includegraphics{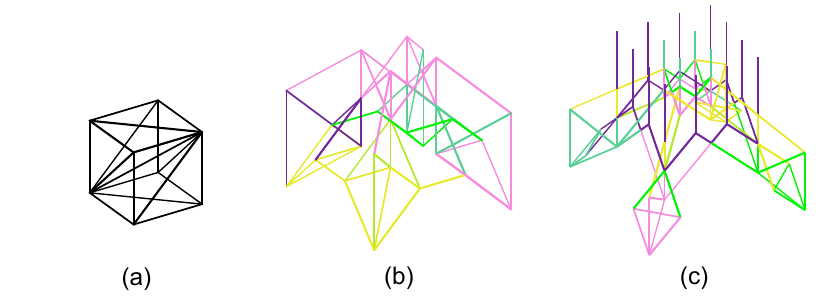}
 \centering
\caption{Test models for the backtracking search algorithm. Model (a) is a simple cube model with one internal crossing and diagonals on every face. Model (b), (c) are topology-optimized vaults with different sparsity \cite{huang2018iass}. Model decomposition is not used for model (a). Model (b) and (c) are decomposed into 8 and 12 groups respectively. Six colors are used cyclically to depict decomposed groups.}
\label{fig::csp_st}
\end{figure}
%
\paragraph{Backtracking search computation statistics}
A computation runtime study is presented in table \ref{table:CSP_overhead} to compare the impact of the heuristic value ordering, model decomposition, and consistency checking on runtime. The backtracking search algorithm is applied to three models shown in figure \ref{fig::csp_st}, with some algorithmic features turned on/off.

From the total runtime shown in table \ref{table:CSP_overhead}, a vanilla backtracking strategy without model decomposition and variable ordering quickly becomes incapable with the increase of element number in the model. With the collision cost turned on but without decomposition, the search algorithm is capable of finding a solution but spends nearly 90\% of its time computing the collision cost. This overhead is caused by the massive collision checking for doing constraint propagation on a large number of variables' domain. With the decomposition turned on, the runtime for computing collision cost quickly drops 97\%, because the decomposition limits the variables' domain that needs to be propagated upon to the ones within the local decomposition group. In addition, the usage of collision cost reduces the number of backtracking and the runtime for testing kinematics, as the collision cost helps prioritizing elements with less chance to cause infeasible kinematics solutions. But this reduction is not sufficient to cover the runtime increase caused by the collision cost's computation.

In contrast, the search algorithm performs the best with the decomposition but without collision cost in all of the three models presented. The winning strategy of the proposed backtracking search algorithm would be using the model decomposition without using the collision cost to dynamically order the values. 

The model (a) (fig\ref{fig::csp_st} (a)) can be viewed as a single layer in a bigger model because of its small number of elements. Even without any backtracking (fig\ref{fig::csp_st} (a) - without collision cost, table \ref{table:CSP_overhead}), the search algorithm spends 7\% of its time checking stiffness\&stability constraint, 65\% checking kinematics, 28\% updating EE direction states. This simple case shows that the computation overhead induced by the geometric and physical constraints. Thus, the extrusion planning problem is dominated by geometrical and physical aspect of the problem, rather than the logical aspect of it. This feature distinguishes this specific class of CSP problem from the common discrete CSP problems considered in the classic AI community, and therefore deserves special algorithmic treatment.

\textbf{Remarks:}
Notice that both updating EE direction states and computing collision cost involves a large amount of two-body collision checking between an element and the end effector extruding along another element. In this work, this two-body collision checker is implemented using an ad-hoc and hand-written line-face intersection to approximate the element-EE collision. In future work, this collision checker can be replaced a more modern collision checker that uses advanced algorithms, e.g. FCL library \cite{fcl2018} with the GJK algorithm \cite{gjk1988} implemented. The authors expect at least 30\% of speed-up on updating EE direction state and computing collision cost if collision checker is replaced.

\paragraph{Routing nodal printing orders}\label{sec:opt_printing_directions}
After the CSP planner finishes its search and produces an extrusion order, the nodal printing orders can be further optimized to increase empirical printing success. For the steps that connect two existing nodes, the assignment of start node and end node can be chosen without affecting the sequence planning result's feasibility. This assignment has recently proven to be critical for the physical execution of spatial extrusion due to the molten joint's incapability to resist bending moment and elastic recoil effect \cite{Gelber_Hurst_Comi_Bhargava_2018}. Gelber et al. introduce a cantilever constraint to their extrusion planning algorithm to address this problem: new elements cannot be connected to node $p$, if any previously printed element connected to $p$ is cantilevered \cite{Gelber_Hurst_Bhargava_2018}. A relaxed version of this constraint is used here to route the nodal printing order: starting from the node with larger degree (number of connected elements) is preferred. Based on the authors' experiments, the introduction of this direction routing process dramatically increases the rate of empirical printing success while not affecting the geometric planning.
 \subsection{Motion planning module}\label{sec:motion_planning_module}
The plan skeleton obtained from the sequence planner specifies the order $O_1, O_2, \dots, O_n$ and a range of collision-free end-effector directions ${\mathbf{v}_{O_i,j}}, \, i \in \{1,\dots,n\} ,\, j \in \{1,\dots,m_i\}$. Notice that this list of vectors corresponds to the directional vectors with boolean constraint variable $V_{O_i,j'} = 0,\, j' \in \{1,\dots,m_V\}$. Their $j'$ indices are renumbered to $j \in \{1,\dots,m_i\}$, where $m_i$ is the total number of collision-free directions for extruding element $O_i$. In this section's description, the vectorial variables are written in bold notation.

To obtain a full motion for the robot, the motion planner needs to (1) determine the robot's trajectory during each extrusion and (2) plan for the robot's trajectory between adjacent extrusions. This is a dual motion planning problem due to the Cartesian motion planning with constraints on end-effector's poses during extrusion task and free motion planning without constraints on end effector's pose in transition. 

\begin{figure}
 \includegraphics[width=1\columnwidth]{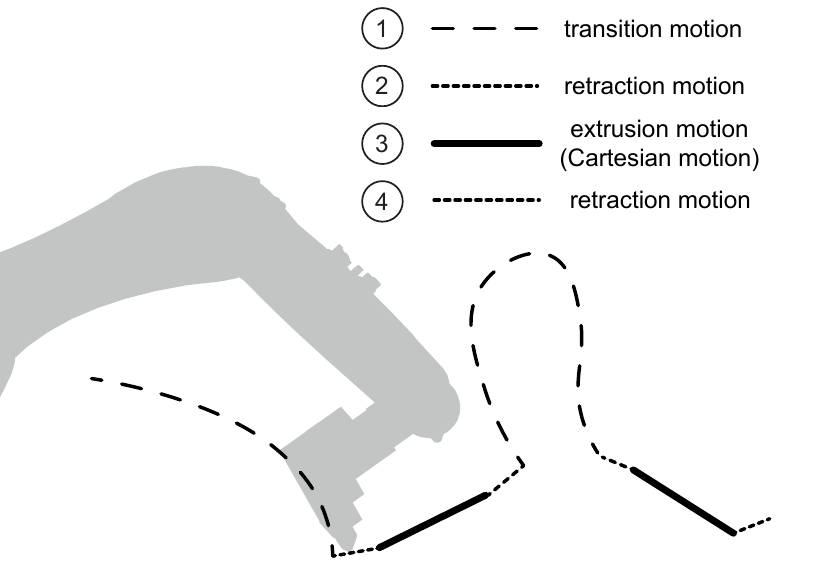}
 \caption{Illustration of a transition motion, retraction motion, and Cartesian motion sequence for two adjacent extrusion processes.}
 \label{fig:hybrid_motion}
\end{figure}

In this work, this dual motion planning problem is solved in two phases: semi-constrained {\it Cartesian planning} (section \ref{sec:semi-constr_cartesian_planning}) to resolve the redundancy in end effector poses and associated robot kinematic during each extrusion task. Then, {\it retraction planning} (section \ref{sec:retraction_planning}) is added between the Cartesian motion and transition motion to enable a safer robot trajectory. Finally, {\it transition planning} (section \ref{sec:transition_planning}) is used to compute robot's trajectory in between adjacent extrusion tasks. The sequential layout of transition motion, retraction motion, and Cartesian motion is illustrated in figure \ref{fig:hybrid_motion}. 

The contribution of this section is a new sampling-based semi-constrained Cartesian planner that is capable to work with long-horizon planning problems. This is a key enabler to make existing motion planning scheme to work with spatial extrusion planning instances, which typically have long planning horizons.
\subsubsection{Semi-constrained Cartesian planning}\label{sec:semi-constr_cartesian_planning}
In many robotic assembly applications, the robot's end effector is required to move linearly, where the end effector's tip must follow designated path points. However, its orientation may still have some degrees of freedom \cite{Maeyer2017descartes}. In the case of spatial extrusion, the tip of the printing nozzle needs to traverse the points on the linear path formed by the element but has freedom in choosing the end effector's orientations. In addition, even when the end-effector's poses are fully specified, there is still kinematic redundancy in choosing corresponding robot configurations. The planning for this type of motion is called semi-constrained Cartesian planning.

In this section, a graph-based semi-constrained Cartesian planner is proposed to resolve the redundancy in the end effector's orientation and robot's kinematics to fully determine robot's joint configuration during each assembly process. For spatial extrusion of a single element, the robot's end effector needs to traverse a linear path with a fixed end effector direction and rotation angle. In order to fully determine the robot's configuration in each individual extrusion task, the planner has three variables to assign: (1) the end-effector direction $\mathbf{v}_k$, (2) the end-effector's rotation angle $r_k$ around its z-axis direction, and (3) joint configurations corresponding to each of the EE poses. As described in section \ref{free_cart_motion}, the EE poses for extruding along an an element are specified by path point Cartesian positions, a direction vector, and a rotation angle.
\begin{centermath}
\begin{aligned}
&\bfp_{k_i} = (x_{k_i}, y_{k_i}, z_{k_i}), \textrm{path points in element } O_k\\
&\bfv_k \in \textrm{feasible direction domain $\mathbf{v}_{O_k,\centerdot}$}\\
&r_k \in [0, 2\pi)
\end{aligned}
\end{centermath}
Solving for robot's kinematic solution problem requires finding feasible joint positions $\mathbf{J}_{k_i}$ for each pose in extrusion task $k$'s path:
\begin{centermath}
\begin{aligned}
&\mathbf{J}_{k_i} = (\theta_{k_i}^{1}, \dots, \theta_{k_i}^{d}), \textrm{d = robot's degrees of freedom}\\
&\mathbf{J}_{k_i} = \textproc{InverseKinematics}(\bfp_{k_i}, \bfv_k, r_k)
\end{aligned}
\end{centermath}
Notice that the inverse kinematics solution $\mathbf{J}_{k_i}$ for target end-effector pose is not unique and needs to be determined by the planning algorithm. Meanwhile, the computed joint solutions have to be collision-free with respect to objects in the environment during the extrusion. In addition, the motion between consecutive joint solutions should respect the robot's maximum velocity and acceleration limitations so that the joint solution sequence is physically executable.

Existing work addresses semi-constrained Cartesian planning problem using an approach that discretizes the end effector's candidate poses and kinematic solutions and performs a discrete search on a planning graph \cite{Maeyer2017descartes}\cite{ROS-I2018Descartes}. This algorithm starts with a list of given end effector poses for the robot to traverse and each end effector pose is assigned with parameters with tolerance ranges. With the tolerance, each given path pose represents a family of parameterized end effector poses and each pose in this family corresponds to a family of robot's joint configurations by performing analytical inverse kinematics with ikfast \cite{ikfast2018}. These joint configurations can be organized as vertices in a planning graph where edges only exist between vertices that belongs to the same or adjacent end effector pose families. Vertices that represent joint configurations in collision are pruned and edges that represents sharp turns of adjacent joint configurations will not be added to the planning graph. A cost is assigned to each edge in the graph as the $L_1$ norm of the difference of the two adjacent joint configurations. In this way, the semi-constrained Cartesian path planning problem is converted to a shortest path searching problem on a directed {\it ladder graph}, which is a multi-partite graph with edge connections between only independent set $k$ and $k+1$, $k \in \{1, \dots, n-1\}$. Each rung in the ladder graph consists of joint configurations that belong to the same end effector pose's parameterized family. The rung's index can be viewed as a time index on path points. The resulting path represents a sequence of joint configurations with minimal joint difference between adjacent joint configurations \cite{Maeyer2017descartes}.

However, the extrusion planning problems usually involve longer planning horizons and two degrees of freedom on choosing EE directional vectors $\mathbf{v}_k$ and rotational angle $r_k$ to determine the EE orientation per extrusion. These features make a direct application of the ladder graph search algorithm described above impractical. For example, for spatial extrusion of a truss model with 300 elements, the storage of the corresponding planning graph will take 362 Gigabytes, which exceeds the RAM capacity of a common desktop computer. To address this memory issue, a sampling-based optimization algorithm is proposed to first search on a sparse representation of the planning graph and then expand this representation into a significantly reduced full graph to perform a shortest path search.
\begin{figure*}[!htbp]
 \includegraphics[width=1\textwidth]{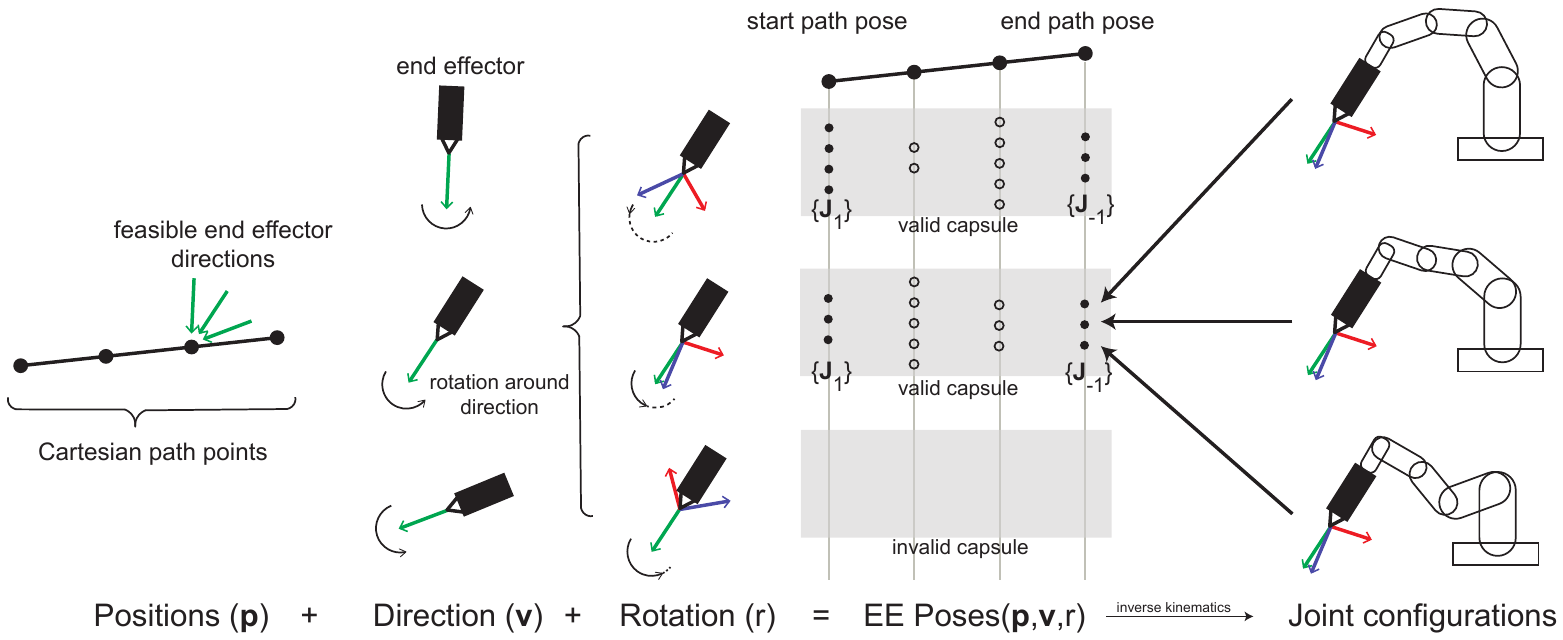}
 \centering
 \caption{Demonstration of capsules in the sparse ladder graph. The gray highlighted fields are capsules. The dots inside a capsule represents the robot's joint kinematics solutions it contains. Inside each capsule, only the start and end pose's kinematic families are stored (black dots), while all other joint poses (white dots) are not stored.}
\label{fig:capsule_illustration}
\end{figure*}
\paragraph{Extracting sparse ladder graph}
This section first introduces a sparse representation of the planning graph, called {\it sparse ladder graph}, to help compress and locate the region on the planning graph that contains a close-to-optimal solution for the semi-constrained Cartesian planning problem. A sampling-based planning algorithm is used to extract this sparse representation that is asymptotically optimal locally in this module, which means that the probability of converging asymptotically to the optimum approaches 1.00 with an infinite number of samples \cite{Karaman_Frazzoli_2011}.

There are two reasons for the memory overhead in the original planning ladder graph: (1) the entire ladder graph needs to be expanded and stored and (2) time indices assigned to workspace path points lead to a massive number of edges connecting rungs that have adjacent time indices. This observation leads to the idea of leveraging the extrusion sequence $\{1,\dots, n\}$ as a sparser time index for rungs and incrementally building a {\it sparse ladder graph} to first find end effector poses for each assembly task and later recover a reduced ladder graph to search for joint configurations. A {\it sparse ladder graph} is a compressed version of the original ladder graph, where joint configurations are grouped under a compact data structure called \textit{capsule} and directed edges are constructed between capsules. 

A capsule is a data structure that contains an extrusion element's index $i$, an EE direction $\mathbf{v}$, an EE rotation angle $r$, and two sets of feasible robot joint configurations for the first and last EE poses specified by the element's path points and orientation $(\mathbf{v}, r)$. The feasible robot joint configuration sets for the first and last extrusion EE pose are denoted by $\{\mathbf{J}_1\}$ and $\{\mathbf{J}_{-1}\}$, respectively. In a word, a capsule is a five-tuple $(i, \mathbf{v}, r, \{\mathbf{J}_1\}, \{\mathbf{J}_{-1}\})$ that stores the information to describe EE poses for an element's extrusion and the robot's feasible joint configurations at the start and the end of the extrusion. A graphical demonstration of capsule's definition is shown in figure \ref{fig:capsule_illustration}.

This definition of capsules enables (1) the definition of cost (or distance) on a directed edge between two capsules based on first and last pose's joint configurations and (2) the later expansion to a full planning graph of joint configurations. Graph edges in the sparse ladder graph are directed and limited to capsules that have adjacent time index, i.e. between $(i, \mathbf{v}_i, r_i, \{\mathbf{J}_1\}_i, \{\mathbf{J}_{-1}\}_i)$ and $(i+1, \mathbf{v}_{i+1}, r_{i+1}, \{\mathbf{J}_1\}_{i+1}, \{\mathbf{J}_{-1}\}_{i+1})$. The cost of such an edge is defined as the minimal $L_1$ norm of joint pose difference between the joint configurations in $\{\mathbf{J}_{-1}\}_i$ and the ones in $\{\mathbf{J}_{1}\}_{i+1}$. By searching on the sparse ladder graph, one can locate close-to-optimal end effector poses for all the extrusion tasks, without encountering the memory overhead caused the expansion of joint configurations for all the path points. 

Notice it is still possible that one can experience memory overflow problem with this sparse ladder graph approach if considering a model with extremely large number of elements. However, the authors find that this sparse ladder graph approach keeps a low memory occupation for all of the experiments that have been carried out and thus the approach is considered sufficient for most of the 3D trusses in the architectural design context.

\paragraph{Computing an optimal capsule path on the sparse ladder graph}
The sparse ladder graph is used to find a path of capsules to traverse all the extrusion tasks that is close-to-optimal locally in this module. This capsule path can be expanded to a fraction of the original planning graph that contains the close-to-optimal path of joint configurations. Sampling-based algorithms are well suited for this problem because they allow an incremental construction of the sparse graph and provide almost-sure convergence to the optimal solutions locally for this module \cite{Karaman_Frazzoli_2011}. 

In order to apply these algorithms to the problem here, special initialization, sampling, feasibility checking, and connecting functions are provided. These procedures adapt planning to the hybrid discrete-continuous state space and the sequential layout of the sparse ladder graph.

Let $X = ([m_1] \times [0, 2\pi)) \times ([m_2] \times [0, 2\pi)) \times \dots \times ([m_n] \times [0, 2\pi))$ be the full state space of the sparse planning problem, where $[m_i] \times [0, 2\pi)$ parameterizes the end-effector's pose by assigning EE direction with index $j \in [m_i] = \{1, \dots, m_i\}$ in a precomputed list of directions and rotation angle $\theta_i \in [0, 2\pi)$. $n$ is the number of elements to be extruded and represents the time index in the extrusion process. For a specific extrusion task $i$, one can consider the task-projected state space $X_i = [m_i] \times [0, 2\pi)$.

A state $x_i$ in the projected state space $X_i$ corresponds to a capsule $(i, \mathbf{v}, r, \{\mathbf{J}_1\}, \{\mathbf{J}_{-1}\})$. Notice that $\{\mathbf{J}_1\}, \{\mathbf{J}_{-1}\}$ are not independent variables, but dependent on on $(i, \mathbf{v}, r)$.

Let $X_i^{obs} \subseteq X_i$ be the set of states where the capsule does not have feasible joint poses for some of the path points for the corresponding task. Let $X_i^{free} = X_i - X_i^{obs}$ be the resulting set of permissible states for extrusion task $i$. Let $\delta: [n] \mapsto X$ be a sequence of states and $\Sigma$ be the set of all paths. The optimal path planning problem on a sparse ladder graph can be defined as the search for the path $\delta^*$ that minimizes the accumulated cost of the path while traversing each extrusion task in a chronological order:
\begin{centermath}
\begin{aligned}
\delta^* = \argmin_{\delta \in \Sigma} \{\sum\limits_{i=1}^{n-1}c(\delta(i),\delta(i+1)) \, | \, \delta(i) = x(i, \cdot, \cdot), \\
\forall i \in [n], \delta(i) \in X_i^{free}\}
\end{aligned}
\end{centermath}
where cost function $c$ is:
\begin{centermath}
\begin{aligned}
&c: X_i \times X_{i+1} \mapsto R_{+}, i \in \{1, \dots, n-1\}\\
&c(x_i, x_{i+1}) = \min || \bfJ_{-1} - \bfJ_1||_{L_1}\\
\textrm{s.t.} &\quad \bfJ_{-1} \in x_i.\{\bfJ_{-1}\}\\
				&\quad \bfJ_1 \in x_{i+1}.\{\bfJ_{1}\}
\end{aligned}
\end{centermath}
%
%
\begin{figure*}[!htbp]
 \centering
 \includegraphics[width=0.9\textwidth]{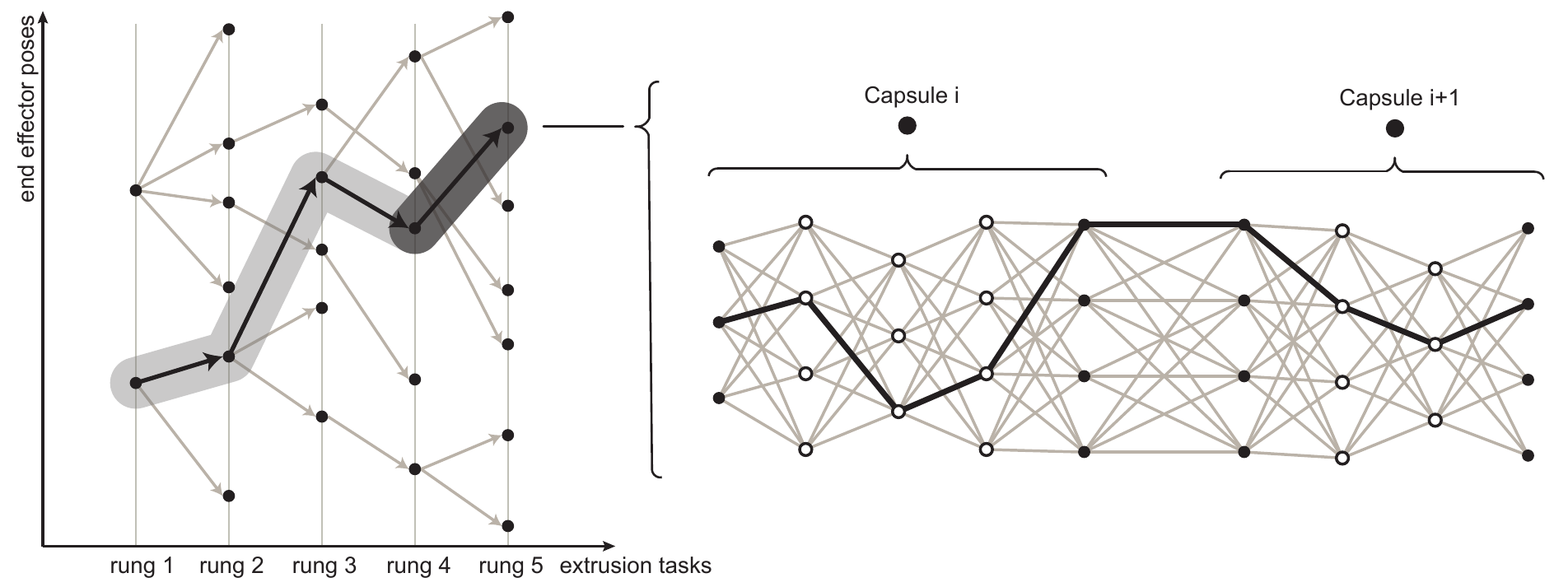}
 \caption{Demonstration on applying RRT* on sparse ladder graph. The optimal capsule path is highlighted, and the expansion of two adjacent capsules is depicted.}
\label{fig:capsule_expansion}
\end{figure*}

In this paper, the Rapidly-exploring Randomize Tree* (RRT*) algorithm is applied to the sparse ladder graph (figure \ref{fig:capsule_expansion}). Other asymptotically optimal sampling-based algorithms, e.g. Probablistic Roadmap* (PRM*), could also be used. The complete description of these algorithms can be found in \cite{Karaman_Frazzoli_2011}. Key modifications enabling these algorithms to operate on the sparse ladder graph are highlighted below: 

\textproc{Sample}: The sampler operates on a hybrid discrete-continuous state space, which returns a state $x_i \in X_i$ that is generated from three different and stand-alone samplers. Each one of these three samplers generates independent samples by randomly sampling uniformly the corresponding variable domain. The generated samples uniquely determine (1) extrusion task's time index $i \in [n]$, (2) end-effector direction index $j \in [m_i]$ in task $i$'s feasible EE direction list and (3) rotation angle $r \in [0, 2\pi)$, which all together determine a state $x_i = (\mathbf{v}_j, r)$.

\textproc{CheckFeasibility}: A state $x_i$'s validity is verified by checking if all the encoded path poses have feasible robot kinematics solutions. A kinematic solution for a given end effector pose is pruned if it results in a collision. Each task has a different set of collision objects, as elements extruded in previous tasks become collision objects in subsequent tasks.

\textproc{Nearest} and \textproc{Rewiring}: Given a state $x_i \in X_i,  1 < i \leq n$, the function \textproc{Nearest} returns the vertex $x_{i-1} \in G \cap X_{i-1}$ with the smallest cost to $x_i$, where $G$ is the current sparse ladder graph. Edge connections are restricted to only vertices in adjacent extrusion tasks, as skipping tasks is not allowed. 
\paragraph{Extracting a trajectory solution}

The sampling-based algorithm returns a path $\delta$ in the sparse ladder graph. The path is then expanded as a subgraph of the original planning graph to enable the use of standard shortest path search algorithms to find the sequence of joint poses with minimal cost. Each state $x_i$ in the returned path $\delta$ is expanded by adding the intermediate path points' kinematic solutions as vertices on the corresponding rungs and then constructing edges between all vertices on adjacent rungs, which corresponds to two successive path points (figure \ref{fig:capsule_expansion}).
 
The expanded graph is a directed acyclic graph (DAG). By topologically sorting its vertices, a shortest path can be identified in time linear in the size of the graph (chapter 24.2, \cite{cormen2009introduction}). The resulting path gives a discretized joint trajectory for each extrusion task in the action sequence that fully determines the robot's configurations during extrusion.

Notice that when applied to semi-constrained Cartesian planning problems, this sparse graph hierarchical approach preserves local optimality in this module asymptotically, compared to directly applying shortest-path search on a full ladder graph of joint configurations. In the original ladder graph, edge connections are limited between joint configurations that belongs to the same or adjacent end effector pose parametrization to satisfy the end effector's orientation constraint. Viewed in the sparse ladder graph, this disallowance of edge connection across pose families is enforced by putting all the capsules that correspond to the same element's extrusion in the same rung (independent set). Thus, no potential decrease in path is lost by applying this hierarchical sparse ladder graph approach locally in this module. However, if viewed globally on the entire extrusion planning system, the joint configurations generated by this module might result in sub-optimal or even infeasible transition trajectories. In general, trading the entire system's completeness and global optimality for tractability is common among hierarchical planning approaches.
\subsubsection{Retraction planning}\label{sec:retraction_planning}
Retraction motion is a short segment of slow linear motion that is inserted between each transition motion and Cartesian motion as a buffer to allow the robot to safely change from high speeds to low speeds when it's approaching (or departing from) the workpiece (figure \ref{fig:hybrid_motion}). In this work, the retraction planner constructs the linear segment by sampling in the set of feasible end-effector's directions produced by the sequence planner and generates a line along this vector with a user-defined length. The same feasibility checking strategy used in the sampling-based algorithm in the last section is applied here to verify the sampled direction's feasibility. Using the Cartesian extrusion motion's orientation, the end effector's orientation during this retraction motion is kept unchanged. 
\begin{figure*}[!htbp]
 \centering
 \includegraphics[width=1\textwidth]{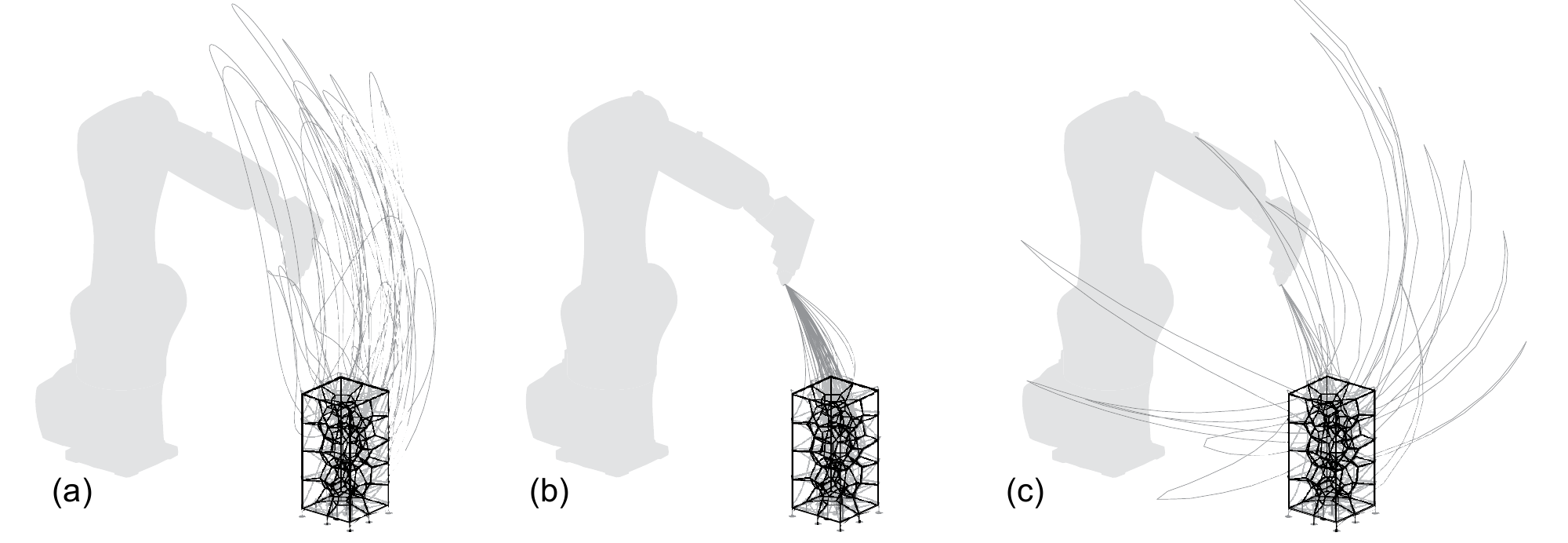}
 \caption{Transition planning with different planners: (a) STOMP \cite{Kalakrishnan_Chitta_Theodorou_Pastor_Schaal_2011} (b) CHOMP \cite{ratliff2009chomp} (c) RRT* \cite{Karaman_Frazzoli_2011}}
\label{fig:transition_planning}
\end{figure*}
\subsubsection{Transition planning}\label{sec:transition_planning}
Following semi-constrained Cartesian planning and retraction planning in the last two sections, transition planning computes a collision-free joint trajectory connecting the last joint pose in the departing retraction motion in extrusion task $i$ and the first joint pose in the approaching retraction motion in extrusion task $i+1$. This is solved using a standard single-query motion planner, which takes into account of the present collision objects in extrusion task $i$ (figure \ref{fig:transition_planning}). The transition planner first tries to call the motion planner for directly connecting the target start and goal configurations. Upon failure, it replans by inserting a reset home waypoint between the start and goal configurations. This guides the planner to find a feasible solution as the configuration space near the home waypoint is less constrained. The transition trajectories generated from three state-of-the-art motion planners are shown in figure \ref{fig:transition_planning}. The result in figure \ref{fig:transition_planning} (b) shows that the CHOMP planner \cite{ratliff2009chomp} frequenly fails to generate a feasible transition plan on its first attempt and thus requires resetting itself to the home waypoint quite often. Based on the authors' experience, the STOMP planner works the best, producing smooth and feasible trajectories with less excessive joint movement.
 \subsection{Post processing module}\label{sec:post_processing_module}
In this work, post processing includes (1) the reassignment of velocities to the computed trajectories and (2) the insertion of end effector control between trajectories. After post processing, the generated commands can be converted into an executable robot code that is specific to an industrial robot's brand. This ``translation'' step is left to external robot's softwares. The post processing module proposed here uses a tagging system to group and tag the trajectories with meta-information that describes the containing process's name and time index. This tagging process enables an easier importing and parsing of the results into various programming systems for application- and hardware-specific adjustment and fine-tuning. This allows the planning framework to be used in various robotic assembly applications with different hardware setups. Two specific ways that the tagging process can be used are described in this section.
\begin{figure*}
 \centering
 \includegraphics[width=1\textwidth]{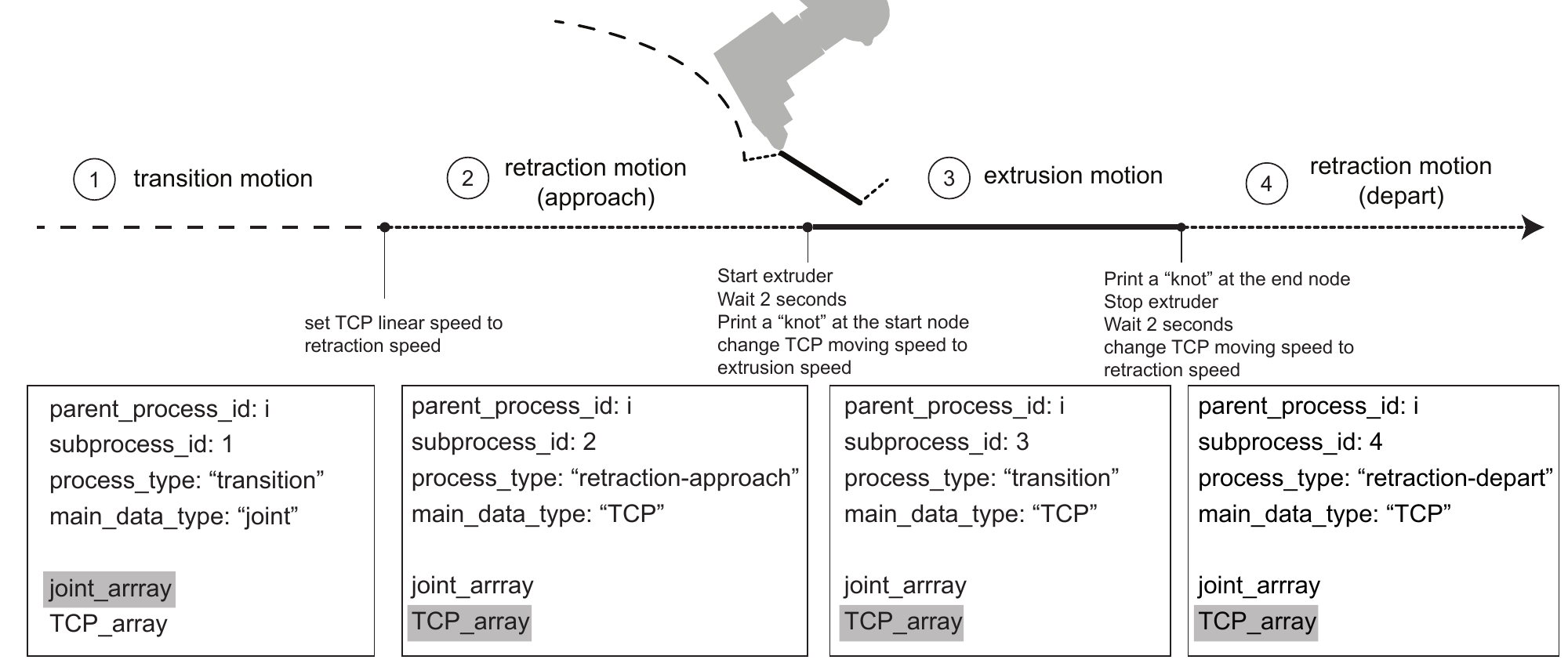}
 \caption{Illustration of the meta-information generated by the tagging system. An element's extrusion process consists of four subprocesses: (1) transition, (2) retraction-approach, (3) extrusion, and (4) retraction-depart. Insertion of end effector control commands and path modifications are shown between processes.}
\label{fig:tagging_system}
\end{figure*}
\paragraph{The reassignment of control velocities and synthesis of IO commands}\label{sec:syn_IO_traj}
The generated robot trajectories is entirely geometrical and its inherent timestamp information only indicates the order of joint configurations. Meaningful timestamps need to be reassigned by the users to the computed trajectories after the planning is finished. In addition, in order to generate instructions for the robot to interact with the physical world, the users need to weave IO commands to synthesize the robot's motion and end effector's behavior. Many existing architectural robotic assembly projects involve an offline programming process. In these projects, the insertion of IO commands is usually carried out in a graphical programming platform, for example, Grasshopper \cite{grasshopper2018}, to have a visually friendly way to insert IO commands in the trajectory command list with live simulation playback. This process, however, can be very tedious when working with robotic assembly applications with a long planning horizon and a massive number of configurations. To increase the computed trajectory's compatibility to these visual programming platforms for trajectory post processing, the generated trajectory is formatted in a customized JSON format, which contains a hierarchical information structure to maximize its readability and usability. Each element's extrusion process contains several subprocesses, each of which is tagged with a subprocess type: transition, retraction-approach, extrusion, or retraction-depart (figure \ref{fig:tagging_system}).

Many robotic assembly projects require the robot to have different end effector speed (also called workspace speed) in different phases of its motion. Users need to produce control velocity subject to the constraint or need of their applications. In the specific case of spatial extrusion, the robot must to extrude material with its end effector following a straight linear movement in a constant speed. Most of the industrial robots provide linear movement commands that take a tool center point (TCP) plane to generate linear movement with a user-defined constant end effector speed. This requires that the result produced by the tagging system contains both robot's joint trajectory and the associated TCP planes to allow users to choose according to the subprocess's definition, i.e. ``switching modes''. To support this feature, when exporting computed trajectories, the planning system performs forward kinematics to every joint configuration to compute corresponding TCP planes. Both of these joint array and TCP array are packed with assembly task id, subprocess id, and subprocess type. In addition, the data type can be specified to indicate what kind of motion the subprocess is using. TCP data should be used if end effector linear movement with constant speed is desired, and joint data should be used if there is no constraint on the end effector's speed.

On the other hand, control commands for the end effector need to be synthesized into the robotic trajectories. These commands are usually application- or hardware-specific, which involves digital IO, analog IO, and wait(idle) times, to enable industrial robot's controller to send commands to activate/stop external customized devices' behavior. For example, spatial extrusion needs the end effector to start extruding material between retraction-approach and extrusion motion, and stop extruding right after the robot finishes the extrusion. This is done by inserting a digital ON/OFF command between the designated processes.

To form a smooth transition into the established method of weaving IO commands in a graphical programming environment, the formatted trajectories produced by the post processor can be imported into any such platform, such as Grasshopper \cite{grasshopper2018}, with a simple customized parser, to decode the JSON file. Users can insert insert robot commands, such as digital IO, analog IO, and wait time, based on the assembly element's index and process context, without having to find the index of a specific joint configuration themselves. Then, existing robot simulation packages can be used to simulate the robot's trajectory to verify the correctness and safety of the trajectories and export brand-specific robot instruction code.
\paragraph{Application-oriented path modification}\label{sec:path_modification}
For many robot assembly processes, especially spatial extrusion, the variety of end effector designs and material properties requires the incorporation of ad-hoc fabrication logic to achieve the desired visual results \cite{hack2014mesh}\cite{helm2015iridescence} or increase the product's structural performance \cite{tam2018}. These fabrication logics, which are derived from physical extrusion experiments, usually involve local modification of an end effector's pose, such as pressing or extruding following small circular movements at structural joints to create local ``knots''.

The metadata associated with the computed trajectories allows users to easily insert these micro path modifications. These path modifications usually require users to iterate on the parameters controlling robot and its end effector's behavior, until they find the best parameter setting based on experimental observations. For spatial extrusion, one needs to perform many experiments to find the delicate balance between the robot end effector's moving speed while extruding, cooling air's pressure, and extrusion rate. Because of the tagging system, the fabrication parameter calibration process repeats between the fine-tuning programming platform and physical tests, while keeping the overall planned robotic trajectory unchanged.

\section{Implementation}\label{sec:implementation}
The proposed hierarchical sequence and motion planning framework has been implemented in a proof-of-concept planning tool called \textit{Choreo}. This tool allows users to compute feasible robotic extrusion trajectories using unconstrained target 3D truss geometry, and it supports customized hardware and work environments. This section first presents the general system architecture (section \ref{sec:impl_system_architecture}) and then presents an overview of the user experience of Choreo along each of its computation state (section \ref{sec:impl_problem_setup} - \ref{sec:impl_post_processing}).
\subsection{System architecture overview}\label{sec:impl_system_architecture}
\begin{figure*}
 \centering
 \includegraphics[width=0.75\textwidth]{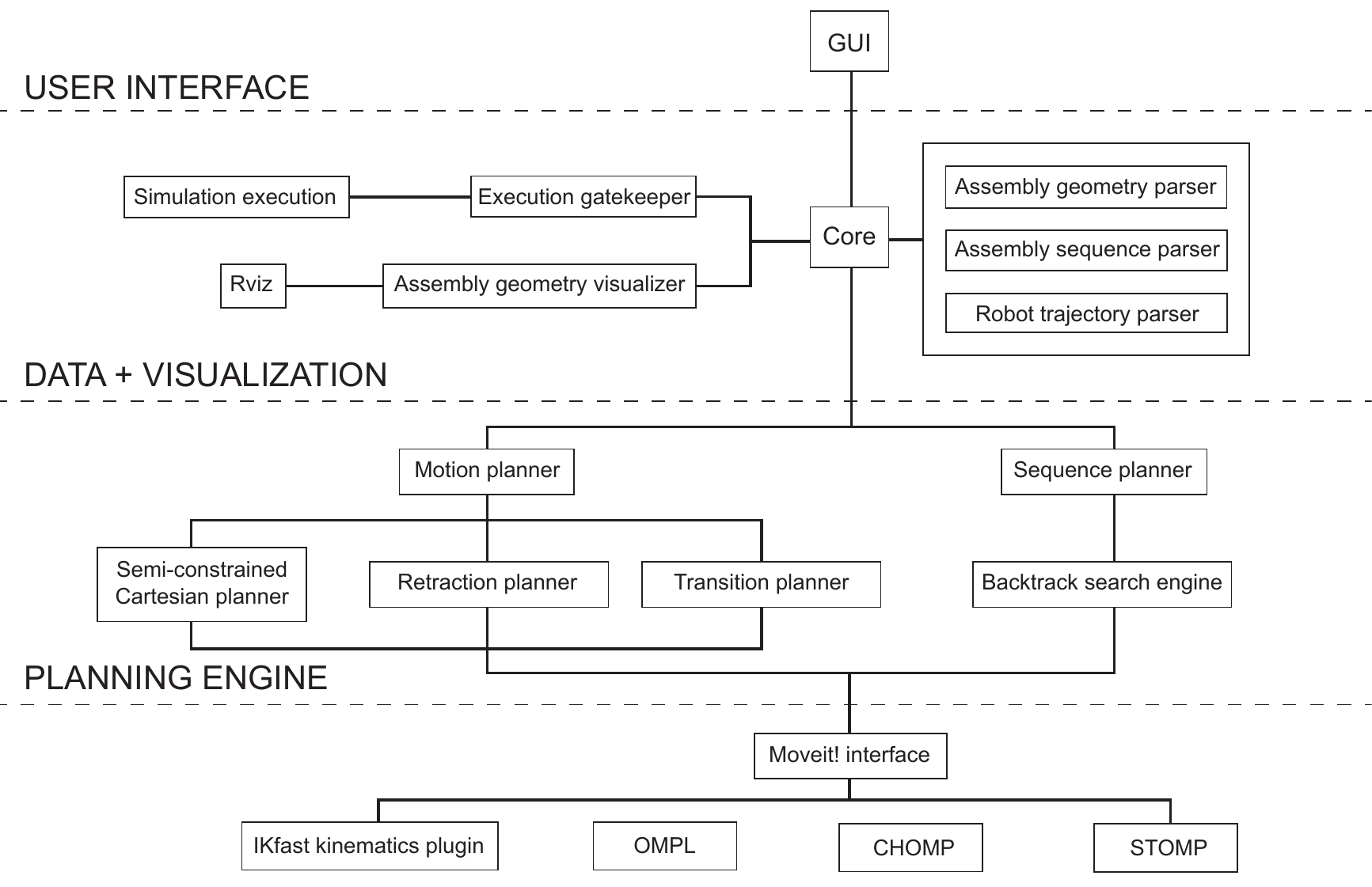}
 \caption{Choreo's system architecture.}
\label{fig:software_system}
\end{figure*}

Choreo is implemented in C++ on the Robot Operating System's (ROS) Kinetic Release on Ubuntu 16.04 \cite{quigley2009ros}. The C++ code is open-source and available online\footnote{\url{https://github.com/yijiangh/choreo}}. Drawing inspiration of the Godel system from ROS industrial \cite{ROS-I2018Godel}, Choreo's system architecture is designed to be modular and flexible: graphical user interface (GUI) module, data IO module, visualization module, and core planning engine modules are all implemented as standalone ROS nodes. Instead of directly communicating to each other, the communication between these modules is coordinated by a central core node using formatted ROS messages and services (figure \ref{fig:software_system}). This enables a clean decoupling between modules that offers users the flexibility to plug in and experiment with their customized sequence or motion planner without changing the rest of the codebase. The GUI is implemented as a simple Qt plugin for the Rviz visualization platform to provide buttons, sliders, and data IO to help users inport and export their data and navigate them through the planning process. 
\subsection{Extrusion problem setup}\label{sec:impl_problem_setup}
Robots and end effectors are specified using a Unified Robot Description Format (URDF) file\footnote{\url{http://wiki.ros.org/urdf}} in Choreo, which is an XML format data that contains robot's link geometry, joint limitation, and other related data. To specify customized end effector, users need to have the STL mesh for the end effector (used for collision checking) and create a URDF file to link the imported geometry to a desired robot link. Static collision objects in the work environment are imported as STL meshes and linked to the robot's URDF file.

The geometry of the 3D trusses can be specified using the node-connectivity format. A decomposition can be added to the geometry model by simply assigning a layer index to each element. The authors develop a simple parser based on the graphical programming platform Grasshopper \cite{grasshopper2018}, to have a visually friendly layer tagging workflow. The relative position between the the build platform and the robot's base can be calibrated from the robot and input into the system by a 3D vector using the GUI widget.
\subsection{Sequence planning}\label{sec:impl_seq_planning}
Currently, Choreo's sequence planner is powered by a customized backtracking search engine (section \ref{sec:solving_CSP}). The analytical kinematics computation is performed through the ikfast kinematics plugin \cite{ikfast2018}. The collision check between robot and the environment is implemented using the collision checking interface provided by Moveit! \cite{sucan2013moveit}. 
\subsection{Motion planning}\label{sec:impl_motion_planning}
The semi-constrained Cartesian planner is implemented based on the Descartes planning package from ROS-Industrial \cite{ROS-I2018Descartes}. The sparse ladder graph and the RRT* algorithm is implemented by the author using the Descartes package's ladder graph data structure. The retraction planner is a direct application of the Descartes package with direction vector sampling.

The transition planner utilizes the motion planner plugin interface of the Moveit! motion planning framework \cite{sucan2013moveit}. Choreo uses the STOMP planner from ROS-industrial \cite{Kalakrishnan_Chitta_Theodorou_Pastor_Schaal_2011}\cite{ROS-I2018IndustrialMoveit} as the main single-query motion planner, but can be easily configured to work other motion planners.
\subsection{Post-processing and execution}\label{sec:impl_post_processing}
\begin{figure*}
 \centering
 \includegraphics[width=1\textwidth]{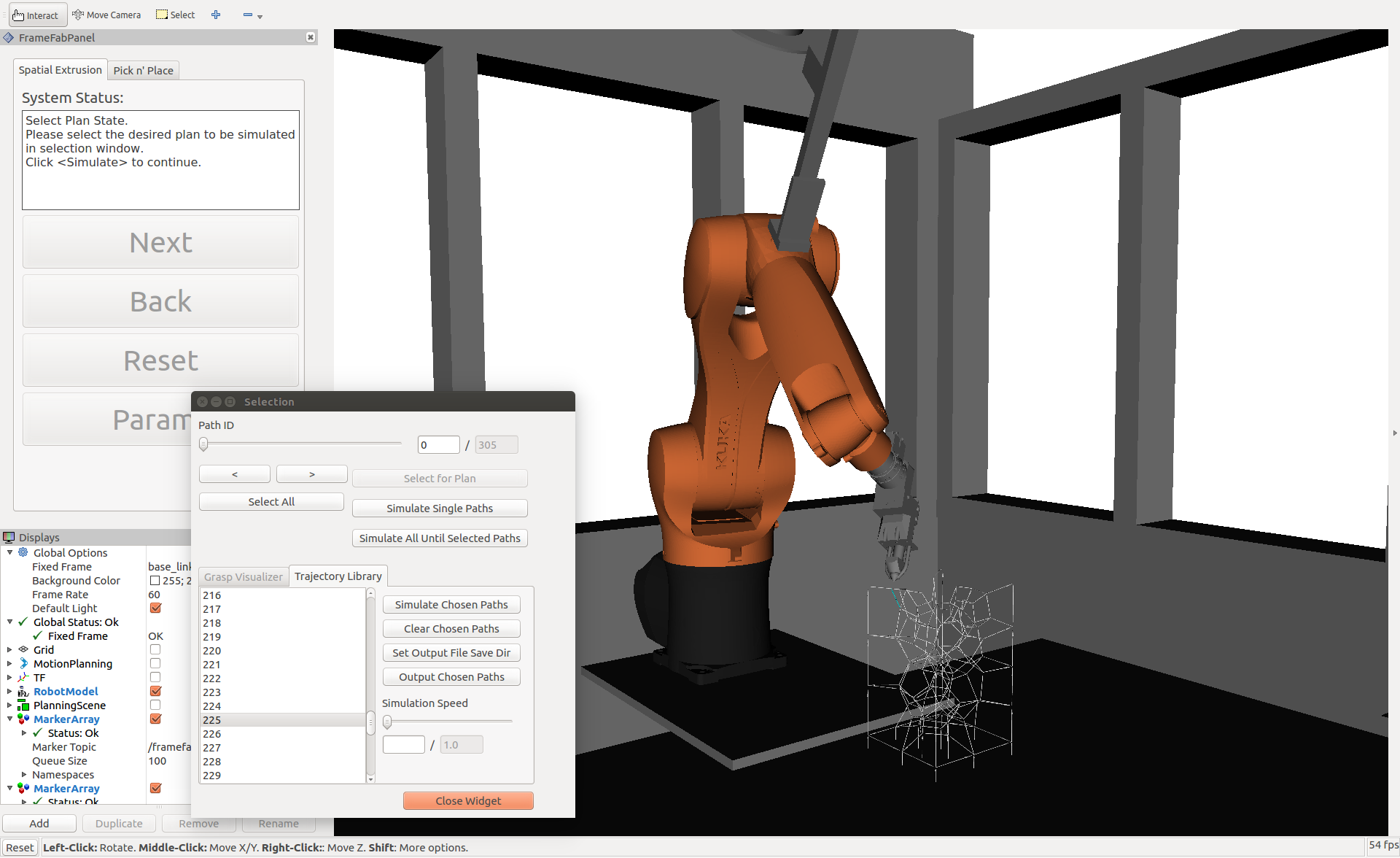}
 \caption{Screenshot of Choreo at trajectory simulation stage.}
\label{fig:rviz_screenshot}
\end{figure*}

After the motion planning is finished, the computed trajectories are tagged with meta-data associated to the extrusion tasks and can be exported as a JSON file. The core module coordinates with the simulation module to display the chosen extrusion tasks' trajectories in Rviz (figure \ref{fig:rviz_screenshot}).

Next, extra post processing and fine-tuning can be performed in other programming platforms. In all the case studies in this work, a customized C\# JSON file parser is implemented in Grasshopper \cite{grasshopper2018}. The KUKA$|$PRC package \cite{braumann2011parametric} and the Robots plugin \cite{Robots2018} are used to post-process the trajectory into a KUKA Robot Language (KRL) file and ABB RAPID file for simulation and execution. The exported trajectory can be configured easily to work in other parametric design platforms and be adapted to other robotic simulation packages such as HAL \cite{schwartz2013hal} and Jeneratiff \cite{dritsas2015digital}. As described in section \ref{sec:robotic_assembly_in_arch}, such simulation tools are useful for visualizing a generated robotic motion plan and generates robotic brand-specific instruction code within the Grasshopper environment.

Hardware-wise, a customized extrusion system was designed and assembled by  Archi-Solution Workshop\footnote{\url{http://www.asworkshop.cn/}}. A detailed description of the end effector, extrusion system, and cooling system can be found in \cite{yu2016acadia} as well as the online supplementary materials of \cite{huang2016framefab}.
 \section{Case Studies}\label{sec:case_studies}
\begin{figure}[h]
 \centering
 \includegraphics[width=1\columnwidth]{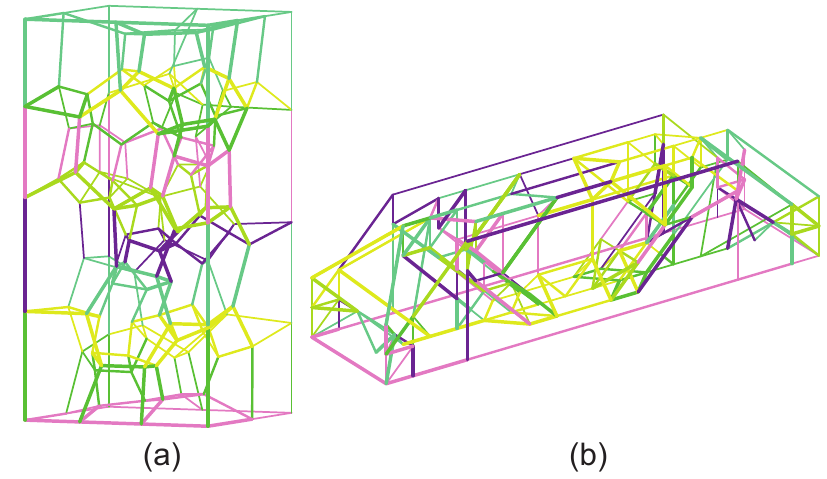}
 \caption{The user-specified decomposition. (a) 3D Voronoi has 10 layers (b) Topopt beam has 53 layers. In the image, six colors are used cyclically to depict layers.}
\label{fig:decomposition_summary}
\end{figure}

To illustrate the capabilities of Choreo, this section describes three case studies that utilize Choreo's power to automatically plan for feasible robotic trajectories for spatial extrusion of complex spatial trusses with non-standard topologies. The presented case studies have fundamentally different topologies and scales: 3D Voronoi (section \ref{sec:voronoi}) and topology optimized simply-supported beam (small and large scale, section \ref{sec:topopt_beam}). Model-related data, together with statistics on assembly planning and fabrication time are presented in table \ref{table:statistics}. The user-guided decomposition of the models are shown in figure \ref{fig:decomposition_summary}. All computational experiments were performed on a Linux virtual machine with 4 processors and 16 GB of RAM on a desktop PC with a quad-core Intel Xeon CPU. Videos illustrating Chore's workflow and the experiments in this paper are available online\footnote{\url{https://www.youtube.com/playlist?list=PLdlQ2M-oI1DxrZu83V7BInOxSFdkqfmKg}}. Additional case studies on robotic extrusion of nonstandard topologies can be found in \cite{huang2018robarch}\cite{huang2018iass}.
\begin{table*}[h]
\centering
  \newcommand{\tabincell}[2]{\begin{tabular}{@{}#1@{}}#2\end{tabular}}
	\begin{tabular}{|c|c|c|c|c|c|c|c|c|c|}
		\hline  Model                  & \tabincell{c}{Node\\count}  & \tabincell{c}{Element\\count}  & \tabincell{c}{Layer\\count} & \tabincell{c}{Sequence\\planning\\time [s]$\dagger$} & \tabincell{c}{S-C Cartesian\\planning\\time [s]} & \tabincell{c}{Retraction\\planning\\time [s]} & \tabincell{c}{Transition\\planning\\time [s]}  & \tabincell{c}{Real\\extrusion\\time [hr]} &\tabincell{c}{Size [mm]}\\
        \hline \tabincell{c}{3D Voronoi\\(sec \ref{sec:voronoi})}
                                       & 148       & 292  & 10  &  2759  & 1253  &10 & 74 (RRT*)   & 3.2 & \tabincell{c}{150\\150\\320}\\
        \hline \tabincell{c}{Topopt beam (small)\\(sec \ref{sec:topopt_beam_s})}
                                       & 121     & 271  	& 53     & 1650    & 1275     & 7 & 667 (STOMP)      & 3.6 & \tabincell{c}{400\\100\\100}\\
        \hline \tabincell{c}{Topopt beam (large)\\(sec \ref{sec:topopt_beam_l})}
                                       & 121     & 271  	& 53     & 1847  &1950  & 160 & 563 (STOMP)        & - & \tabincell{c}{2800\\700\\700}\\
        \hline
	\end{tabular}
	\caption{Input model information, computation statistics, and fabrication time of the case studies. Layer count is the number of layers used in the user-generated decomposition (figure \ref{fig:decomposition_summary}). S-C Cartesian planning represents semi-constrained Cartesian planning. Size is shown in a length-width-height format. $\dagger$ The backtracking algorithm does not use value ordering but incorporates decomposition and constraint propagation.}
	\label{table:statistics}
\end{table*}
\subsection{3D Voronoi}\label{sec:voronoi}
\begin{figure*}[h]
 \centering
 \includegraphics[width=1\textwidth]{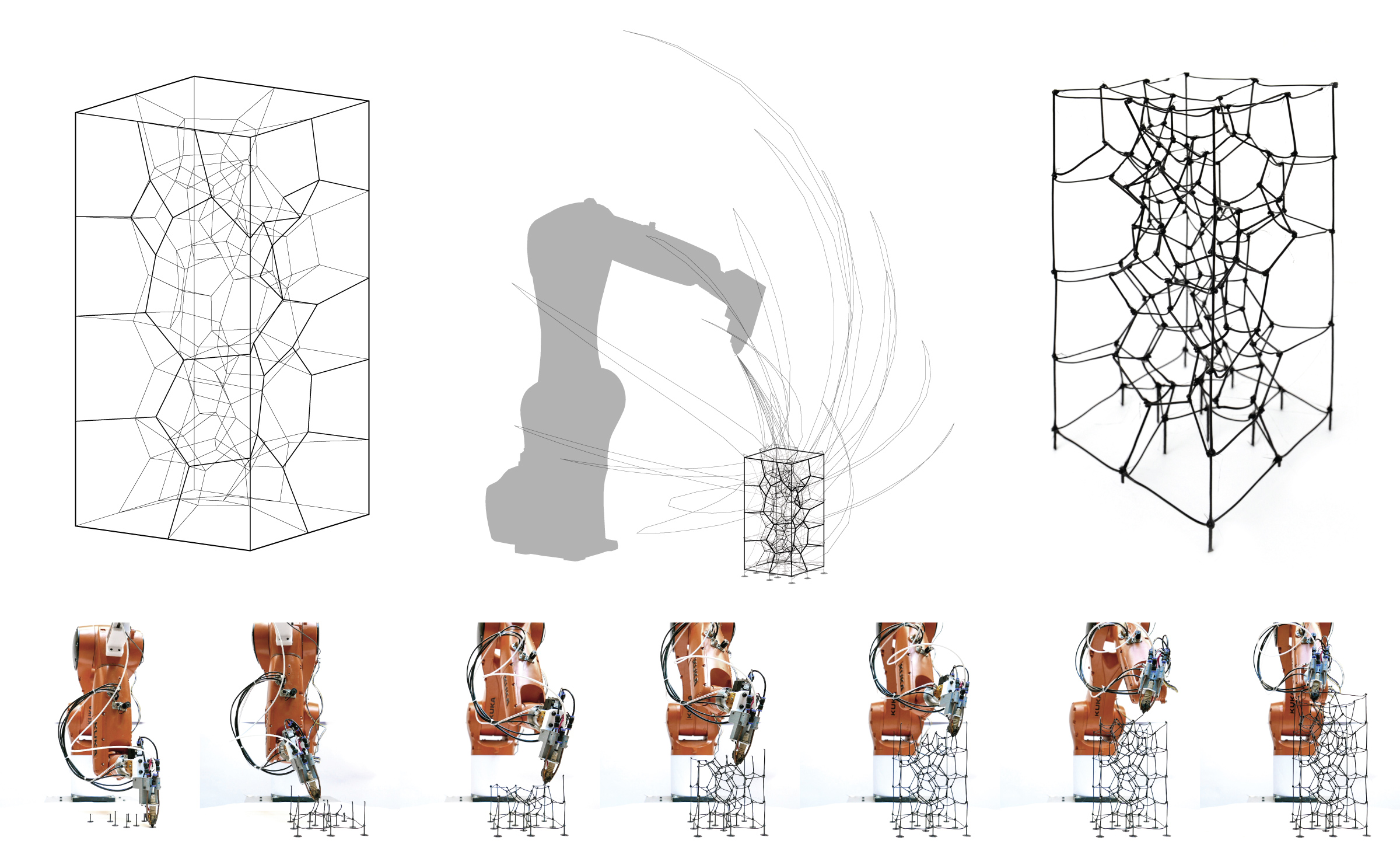}
 \caption{3D Voronoi design, robotic trajectories with RRT*, and final extruded result. A fixed-base KUKA KR6-R900 robotic arm is used.}
\label{fig:voronoi_result}
\end{figure*}

The 3D Voronoi design was generated by randomly sampling points within a rectangular solid, and then using the 3D Voronoi component in Grasshopper \cite{grasshopper2018} together with Kangaroo2 \cite{kangaroo2018} to initiate the 3D Voronoi pattern. A sphere collision algorithm was used to force the element lengths to have low variance. A KUKA KR6-R900 robot was used to execute the extrusion. Figure \ref{fig:voronoi_result} shows the design and fabrication of this structure. 
Because of the Voronoi-generating algorithm, there is low variation in node valence, and most nodes only connect four elements. Elements are well supported during each construction step, and there are few very long elements. The internal topology does not have a trivial layer-based pattern. Thus, it is unintuitive for humans to find a sequence manually, and the Choreo platform proves helpful.

However, elements at the boundary have smaller node valences and very long length. Even though the geometrically planned trajectory is feasible, the extruded element deviates from its ideal position because of the material's thermal wrapping. This deviation is sometimes large enough that the robot is not able to connect to these elements in the subsequence extrusion processes by following the computed trajectories. This issue is resolved by adding micro-path modifications to the computed Cartesian extrusion path in the post processing stage to extrude a ``knot'' at the node to compensate for the inaccuracy brought by the thermal behavior of the material.
\subsection{Topology optimized simply-supported beam}\label{sec:topopt_beam}
Using the ground structure topology optimization method described in \cite{huang2018iass}, a simply-supported beam was designed for the loads and boundary conditions shown in figure \ref{fig:beam_result} (a), (b). The resulting topology is fairly irregular when compared to a standardized mesh topology. The beam is scaled to a small size and a large size and two different machine setups are used in the planning. The large-scale example is presented to demonstrate the potential of applying Choreo at the scale of a real building component, which in particular fits into the context of construction robotics.

\subsubsection{Small scale}\label{sec:topopt_beam_s}
\begin{figure*}[h]
 \centering
 \includegraphics[width=1\textwidth]{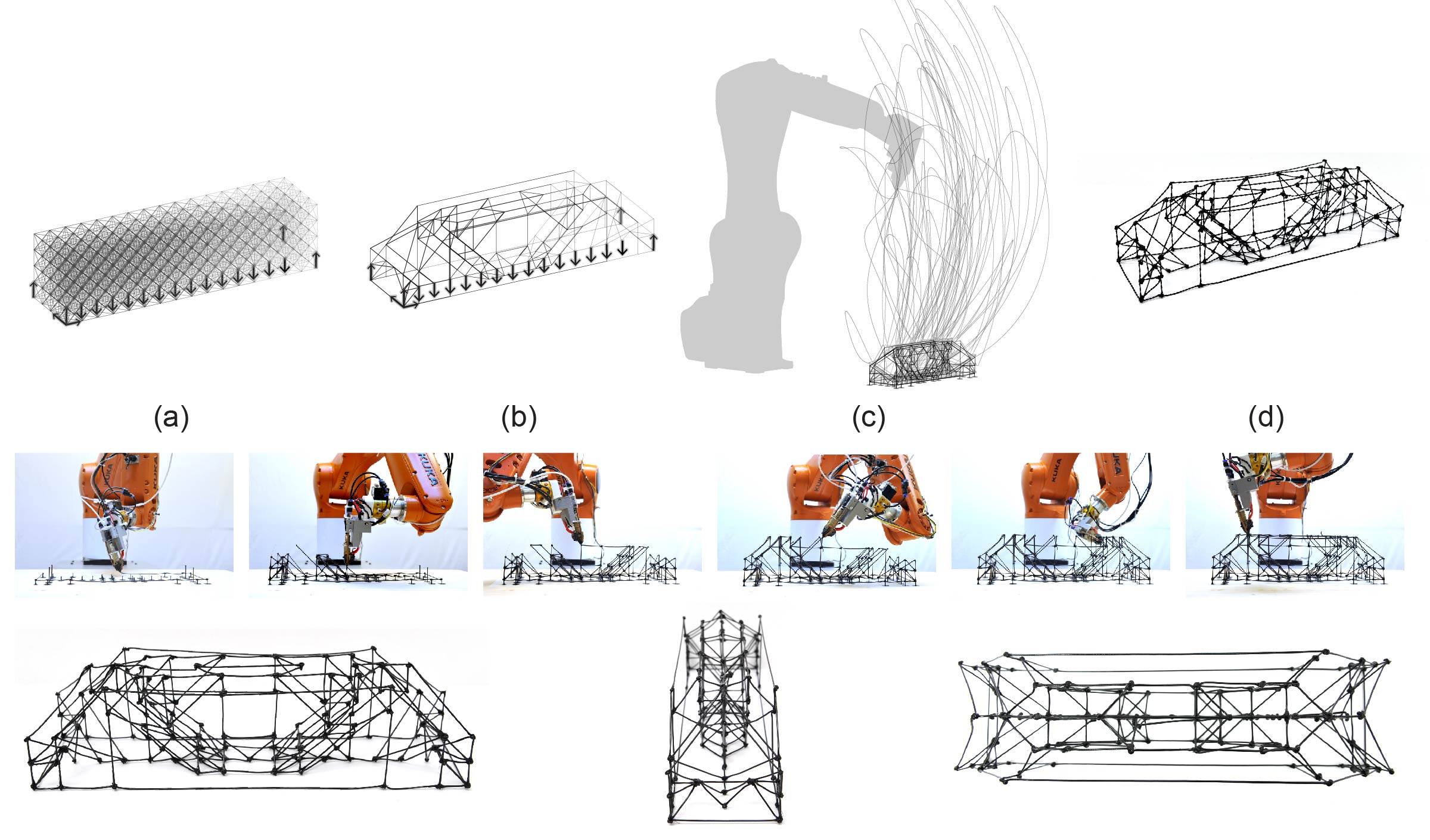}
 \caption{Topology optimized simply-supported beam, with (a-b) topology optimization input and result, (c) robot trajectories with STOMP, and (d) final extruded result. A fixed-base KUKA KR6-R900 robot is used.}
\label{fig:beam_result}
\end{figure*}
The small-scale beam spans 400 millimeters. A fixed-base KUKA KR6-R900 (maximal reach 0.9m) robot is used to execute the extrusion. The average element length of the model is fairly long, and element length variation is low because the design is generated from a regular base mesh. However, the geometric configurations generated from these elements is not trivial. The trajectory highlighted in figure \ref{fig:beam_result} shows the corresponding tool center point traveling trajectory from the transition planning result, indicating that the robot's configuration changes significantly between many pairs of adjacent extrusions. As a result, trajectories that respect joint limits and avoid collisions are long and unintuitive to humans.
\subsubsection{Large scale}\label{sec:topopt_beam_l}
\begin{figure}[h]
 \centering
 \includegraphics[width=1\columnwidth]{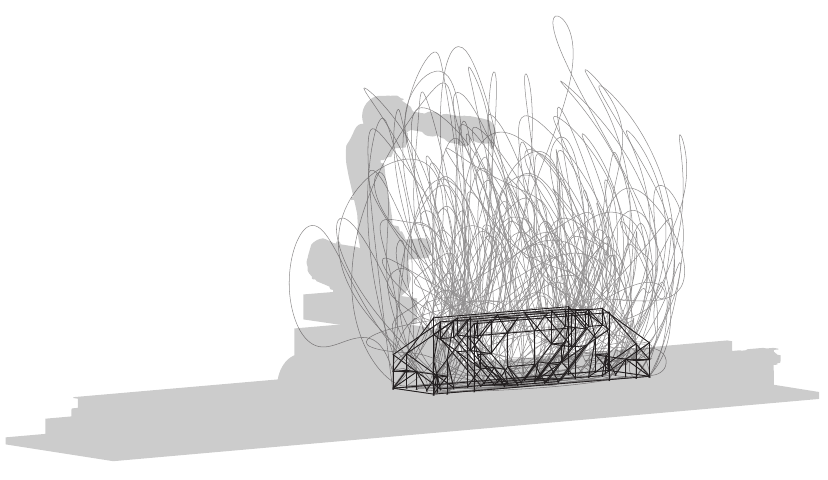}
 \caption{Simulated end-effector trajectory of an ABB IRB6640-185-280 mounted on a 5.4-meter linear track to spatially extrude of the large-scale topology optimized beam.}
\label{fig:L_beam_sim_result}
\end{figure}

The large-scale beam has a span of 2.8 meters, with 0.7 meters in thickness and height. A 5.4-meter linear track is added to an ABB IRB6640-180-280 robotic arm (maximal reach 2.8m) to accommodate the scale of the beam. The additional degree of freedom from the linear track expands the feasible workspace of the robot. Specifically, this extra dimension increases the set of kinematic solutions, making it easier to find a feasible joint configuration. In practice, the movement of the six revolutional axes of the robotic arm is preferred over the track's movement for accuracy and energy consumption reason. In this paper, in order to to penalize extra movement of the prismatic linear joint, an extra penalty weighting factor is added to linear track's joint movement when computing the $L_2$ distance between joint configurations for constructing ladder graph in the semi-constrained Cartesian planning module (section \ref{sec:semi-constr_cartesian_planning}). The analytical inverse kinematics of the robot is done by discretizing the prismatic joint and attempting 6-dof IK. This discretization resolution also balances the completeness and RAM overhead of the computation and could be iteratively increased to find a feasible solution or limit excessive base movement. In this experiment, the prismatic joint is discretized every 0.01 m over a full length 5.4 m. A more in-depth investigation of the robot base placement, using reachability analysis \cite{Makhal_Goins_2017}, is under investigation for future work. The resulting simulated extrusion trajectory is shown in figure \ref{fig:L_beam_sim_result}.
 \section{Conclusion}\label{sec:conclusion}
\subsection{Summary of contributions}
This paper presents the first attempt to rigorously formulate the sequence and motion planning (SAMP) problem in the architectural robotic extrusion context using a CSP encoding. The research presented provides the first integrated algorithmic solution that uses a new hierarchical planning algorithm to tackle the extrusion SAMP problems with long planning horizons and complex geometry. Along with the planning algorithm, this paper also presents empirical computation results to provide insight into the communicational bottleneck that differentiates extrusion TAMP from the general discrete CSP problems. The presented algorithm and data serves as a comparing baseline for future algorithmic development.

This paper also presents a proof-of-concept software implementation, called Choreo, which is a hardware- and application-agnostic. Choreo can be easily configured to work with industrial robots across brands, sizes, and joint spaces. Three case studies involving spatial trusses with non-standard topologies, including two real fabrication experiments and one large-scale assembly simulation, are presented to demonstrate the new fabrication possibilities enabled by applying the proposed planning framework.
\subsubsection{Potential impact}
The case studies presented in this work have demonstrated how the proposed extrusion planning framework's integration into existing digital design workflow can support topology as a fundamental design variable on designers' palette for robotic spatial extrusion. The emergence of this type of high-level automated planning system can provide a better way for designers to interact with robots, shifting the machine programming experience back to high-level tasks in the architectural language of shape and topology.

On the other hand, the flexibility and the efficiency of Choreo creates a testbed for educators, researchers, and practitioners to explore the fabrication of 3D trusses across different topological and scale classes more boldly. It provides a general common ground for future research in extrusion-oriented SAMP planner, and creates a bridge between architectural robotics research community and task and motion planning research community.
\subsubsection{Limitation and future work}
One key limitation of the current work is on the need of human intervention for model decomposition to accelerate the sequence planner. An automatic decomposition algorithm will eliminate this last bit of human intervention and fully automate the planning process.

Potential directions of future investigation are summarized as follows:
\paragraph{Backtrack between planning layers}
When the planning system encounters a planning failure in any layer in the hierarchy, there is no backtracking mechanism provided to allow it to backtrack across planning layers. Thus, the hierarchical planning algorithm is not algorithmically complete for the entire system, i.e. guaranteed to find a solution if one exists, although the algorithm proposed in each level of the hierarchy is complete. Existing work in TAMP has devised various way to allow this geometric backtracking across planning layers. The integration of some of this research is left as future work.
\paragraph{Extension to other robotic assembly applications}
All the algorithmic descriptions and case studies presented in this work are performed in the context of the specific application of robotic spatial extrusion. Spatial printing of 3D trusses can be viewed as assembling linear plastic beam elements in the space with melted plastic joint connection. Generalizing the proposed planning framework to a broader class of assembly applications, such as spatial positioning, requires little modification of the algorithm. These modifications include (1) different predefined plan skeletons, (2) different constraints on the end effector's orientation in the semi-constrained Cartesian planner that takes the geometry of the end effector, element, and connection detail between elements into consideration. However, notice that this extension to spatial positioning applications does not include the automatic process for joining elements, but only include the sequence and motion planning for a sequence of pick - transition - place - transition motions, assuming an external process handles the joining process flawlessly. 
\subsubsection{Concluding remark}
Automatic sequence and motion planning (SAMP) has been a key missing link in the established digital design-robotic spatial extrusion workflow. Architectural robotic extrusion SAMP problems post unique technical challenges, including (1) long planning horizon and (2) structural stiffness, stability and geometric collision constraints for extrusion sequence planning. This paper presents a new algorithmic framework and a software tool called Choreo to fill in this missing link. The presented research clears away the technical barrier of the sequence and motion planning that has been congesting the workflow and limiting design-build freedom. Although Choreo is still in its early stage, its flexibility and speed have already suggested an exciting future possibility: fabrication and assembly logic related to robotic constructibility could be integrated as a driver in iterative conceptual design, pushing the role of technical assessment from checking a nearly finalized design to an early-stage design aid.

%

\begin{acknowledgement}
The authors want to acknowledge Thomas Cook,
Khanh Nguyen, and Kodiak Brush at MIT for their work on constructing
the early-stage prototype of the software and hardware presented in
this work. The authors would also like to thank Jonathan Meyer from
ROS-Industrial for his insightful discussion on GitHub, Zhongyuan
Liu from University of Science and Technology of China for his help
on generating the 3D Voronoi shape, Lavender Tessmer and Zhujing
Zhang at MIT for their help on diagram drawing. The authors want to
thank Archi-Solution Workshop (http://www.asworkshop.cn/) for their
support on the designing and assembling of the mechanical extrusion
system used in the case studies. Caelan Garrett acknowledges the support
from NSF grants 1420316, 1523767 and 1723381, from AFOSR
FA9550-17-1-0165, from ONR grant N00014-14-1-0486, and an NSF
GRFP fellow-ship with primary award number 1122374. Any opinions,
findings, and conclusions or recommendations expressed in this material
are those of the authors and do not necessarily reflect the views
of the sponsors.
\end{acknowledgement}

\bibliographystyle{spmpsci}

\end{document}